\pdfoutput=1
\documentclass[11pt]{article}

\usepackage[final]{acl}

\usepackage{times}
\usepackage{latexsym}
\usepackage{graphicx}
\usepackage{hyperref}
\usepackage{url}
\usepackage{subfigure}
\usepackage{wrapfig}
\usepackage{booktabs}
\usepackage{multirow}
\usepackage{caption}
\usepackage{tabularx}
\usepackage{amsmath}
\usepackage{amsfonts}
\usepackage{tablefootnote}
\usepackage {stfloats}
\usepackage[T1]{fontenc}
\usepackage{listings}
\usepackage[utf8]{inputenc}

\usepackage{microtype}

\usepackage{inconsolata}
\title{SMR: State Memory Replay for Long Sequence Modeling}

\author{
  Biqing Qi\textsuperscript{1,2,4,}$^*$, 
  Junqi Gao\textsuperscript{3,}\thanks{Equal contributions.}, 
  Kaiyan Zhang\textsuperscript{2},
  Dong Li\textsuperscript{3},
  Jianxing Liu\textsuperscript{1},  
  Ligang Wu\textsuperscript{1,}\thanks{Corresponding authors: Bowen Zhou and Ligang Wu.}, 
  Bowen Zhou\textsuperscript{2,}$^\dag$ \\
  $^1$ Department of Control Science and Engineering, Harbin Institute of Technology, \\
  $^2$ Department of Electronic Engineering, Tsinghua University, \\
  $^3$ School of Mathematics, Harbin Institute of Technology, \\
  $^4$ Frontis.AI, Beijing \\
  {\tt\small \{qibiqing7,gjunqi97,arvinlee826\}@gmail.com,} \\ 
  \tt\small {zhang-ky22@mails.tsinghua.edu.cn},
  {\tt\small \{jx.liu,ligangwu\}@hit.edu.cn, zhoubowen@tsinghua.edu.cn}
  }
 
\begin{document}

\maketitle

\begin{abstract}
Despite the promising performance of state space models (SSMs) in long sequence modeling, limitations still exist. Advanced SSMs like S5 and S6 (Mamba) in addressing non-uniform sampling, their recursive structures impede efficient SSM computation via convolution. To overcome compatibility limitations in parallel convolutional computation, this paper proposes a novel non-recursive non-uniform sample processing strategy. Theoretical analysis of SSMs through the lens of Event-Triggered Control (ETC) theory reveals the Non-Stable State (NSS) problem, where deviations from sampling point requirements lead to error transmission and accumulation, causing the divergence of the SSM's hidden state. Our analysis further reveals that adjustments of input sequences with early memories can mitigate the NSS problem, achieving Sampling Step Adaptation (SSA).
Building on this insight, we introduce a simple yet effective plug-and-play mechanism, State Memory Replay (SMR), which utilizes learnable memories to adjust the current state with multi-step information for generalization at sampling points different from those in the training data. This enables SSMs to stably model varying sampling points. Experiments on long-range modeling tasks in autoregressive language modeling and Long Range Arena demonstrate the general effectiveness of the SMR mechanism for a series of SSM models.

\end{abstract}

\section{Introduction}
    Long sequence modeling has attracted extensive interest due to its broad prospects in natural language processing \citep{beltagy2020longformer,BrownMRSKDNSSAA20,Ouyang0JAWMZASR22}. The mainstream architectures for sequence modeling mainly focus on attention-based Transformers \citep{vaswani2017attention}. However, the quadratic complexity of softmax attention brings a computational bottleneck \citep{choromanski2020rethinking,wang2020linformer,beltagy2020longformer}, which makes attention-based architectures inefficient for handling long sequences. Although the introduction of linear attention \citep{wang2020linformer} reduces the computational complexity, it cannot well approximate the performance of the vanilla Transformer. More importantly, purely attention-based architectures cannot capture long-range dependencies well. On the other hand, state space model (SSM)-based architectures \citep{gu2021efficiently,gupta2022diagonal} show superior performance on the Long Range Arena (LRA) \citep{Tay0ASBPRYRM21}  benchmark for long sequence modeling due to their linear computational complexity and excellent long-range dependency capturing ability.
    
    Existing SSM-based model architectures, such as S5 \citep{SmithWL23} and S6 \citep{abs-2312-00752}, primarily rely on recursive structures to tackle the varying sampling step issue. S5 introduced learnable step sizes for each step to improve the Sampling Step Adaptation (SSA) capability of SSM. S6 (Mamba) introduced data-dependent parameter settings, which makes the state propagation of the SSM model more flexible. 
    However, this restricts its inference computation to parallel scanning instead of the original efficient convolution mode, significantly hampering training efficiency and imposing a heavier inference burden when handling long inputs at once.
    
    To address the mentioned issues, we aim to propose a method that goes beyond recursive constraints to improve SSA capability. This strategy seeks to enhance the SSM, making it more adaptable and flexible for various parallel convolution computation types, including advanced architectures like S4 \citep{GuGR22}, Mega \citep{MaZKHGNMZ23}, SPADE \citep{abs-2212-08136}, and more.
    Specifically, we leverage the Event-Triggered Control (ETC) Theory \citep{HeemelsJT12,Tabuada07} to provide the first demonstration of the Non-Stable State (NSS) problem in SSMs. We show that for a fixed-parameter SSM, varying sampling steps, deviating from the model's sampling point requirements, triggers error propagation and accumulation, ultimately leading to the divergence of the hidden state. Our analysis further reveals that adjustments based on early memories of the input sequence can achieve SSA, effectively solving the NSS problem. Inspired by this finding, we propose a simple yet effective plug-and-play mechanism, State Memory Replay (SMR), it can significantly alleviate the NSS problem in SSMs by improving SSM the capability of SSA thus bring further sequence modeling capabilities. 
    In particular, SMR can achieve better generalization ability at different sampling points, especially when dealing with the stochastic selected sampling points. 
    We conduct experiments on autoregressive language modeling on Wikitext-103 \citep{MerityX0S17} and long sequence modeling on LRA. The results show that the SMR mechanism can bring better performance to SSM-based model, on both autoregressive language modeling and long sequence modeling tasks. It can also further improve a series of competitive SSM-based models such as S5, SPADE, Mega, and S6, which verifies the generality and effectiveness of the proposed SMR mechanism.
    In summary, our main contributions are three folds:
    \begin{itemize}
        \item We are the first to identify the NSS issue in SSMs. We theoretically analyze and experimentally verify the issue from a novel perspective of ETC theory, demonstrating that inputs that do not satisfy the stability condition can lead to the divergence of the hidden states of SSMs and affect model performance.        
        \item Based on our theoretical analysis and experimental results, we reveal that adjustment of the input sequence with early memory can achieve adaptive sampling adjustment capability to solve the NSS problem. Motivated by this, we propose the SMR mechanism.
        \item SMR is able to enhance the existing SSM series models to improve sampling point generalization and sequence modeling capabilities in some real-world tasks with varying design sampling points, including autoregressive language modeling and long sequence modeling, without affecting computational efficiency.
    \end{itemize}
\section{Preliminaries: State Space Models}
The state space model is formally defined by eq.(\ref{eq1}) and eq.(\ref{eq2}):
\begin{equation}
\label{eq1}
\dot{\boldsymbol{x}}(t) ={\boldsymbol{A}} \boldsymbol{x}(t)+{\boldsymbol{B}} {u}(t),
\end{equation}
\begin{equation}
\label{eq2}
{y}(t) = {\boldsymbol{C}} \boldsymbol{x}(t)+ {\boldsymbol{D}} {u}(t),
\end{equation}

where $\boldsymbol{A}\in \mathbb R^{n \times n}$, $\boldsymbol{B} \in \mathbb R^{n \times m}$, $\boldsymbol{C}\in \mathbb R^{m\times n}$, $\boldsymbol{D}\in \mathbb R^{m\times m}$, ${u}(\cdot):\mathbb R\mapsto \mathbb R^m$ denotes the input sequence with dimension $m$, and $\boldsymbol{x}(\cdot):\mathbb R\mapsto \mathbb R^n$ is the latent state.

\paragraph{S4}Previous works \citep{GuJGSDRR21, GuGR22} formed the S4 model, which constructed a set of Structured State-Space Sequence Model (S4) parameters for each dimension of the input $u$ to construct an Single-Input, Single-Output (SISO) system, i.e., for an input $u:\mathbb{R} \rightarrow \mathbb{R}^m$, the same set of SSM parameters is broadcasted to each dimension $u^{(p)}: \mathbb{R} \rightarrow \mathbb{R}$. Specifically, they employed the bilinear method to perform discretization:
\begin{align}
\label{eq3}
\boldsymbol{x}_{k}=\overline{\boldsymbol{A}} \boldsymbol{x}_{k-1}+\overline{\boldsymbol{B}} {u}_{k}^{(p)},  \\
{y}_{k}=\overline{\boldsymbol{C}} \boldsymbol{x}_{k} ,
\end{align}
where $\overline{\boldsymbol{A}}=(\boldsymbol{I}-\Delta t / 2 \cdot \boldsymbol{A})^{-1}(\boldsymbol{I}+\Delta t / 2 \cdot \boldsymbol{A})$, $ \overline{\boldsymbol{B}}=(\boldsymbol{I}-\Delta t / 2 \cdot \boldsymbol{A})^{-1} \Delta \boldsymbol{B}\in\mathbb R^{n\times 1}$, $\overline{\boldsymbol{C}}=\boldsymbol{C}\in\mathbb R^{n\times 1}$. The matrix $D$ is omitted here because it can be viewed as a residual connection. For each element $u^{(p)}, p\in{1, 2, \dots m}$, $t$ is a fixed discretization step, the same for each step. Then, the S4 became a parameterized model with trainable parameters
$\overline{\boldsymbol{A}}$, $\overline{\boldsymbol{B}}$, $\overline{\boldsymbol{C}}$, and $\Delta t$. By assuming $x_0 = \mathbf 0$, we can obtain:
\begin{equation}
\label{eq5}
{y}_{k} =\overline{\boldsymbol{C A}}^{k-1} \overline{\boldsymbol{B}} {u}_{1}+\cdots+\overline{\boldsymbol{C A}} \overline{\boldsymbol{B}}u_{k-1}+\overline{\boldsymbol{C B}} {u}_{k},    
\end{equation}
thus the output could be calculated efficiently by convolution $y =\overline{\boldsymbol{K}} * {u}$, where
\begin{equation}
\label{eq6}
\begin{aligned}
&\overline{\boldsymbol{K}} \in \mathbb{R}^{L}:= \mathcal{K}_{L}(\overline{\boldsymbol{A}}, \overline{\boldsymbol{B}}, \overline{\boldsymbol{C}}):=\left(\overline{\boldsymbol{C A}}^{i} \overline{\boldsymbol{B}}\right)_{i \in[L-1]}\\
&=\left(\overline{\boldsymbol{C B}}, \overline{\boldsymbol{C A B}}, \ldots, \overline{\boldsymbol{C A}}^{L-1} \overline{\boldsymbol{B}}\right),
\end{aligned}
\end{equation}
is the convolution kernel and $L$ is the sequence length. With their proposed Normal Plus Low-Rank (NPLR) parameterization, the S4 Convolution could be calculated in $\widetilde{O}(L+m)$ operations.
\paragraph{S5}Given the uniform time step employed by S4 for each time interval, it encounters difficulties when confronted with irregularly sampled data. To overcome this limitation, S5 \citep{SmithWL23} introduced adaptive and learnable step sizes for each time step, enhanced its capability to effectively handle irregularly sampled data. Furthermore, S5 extended the S4-established Single-Input Single-Output (SISO) system to a more versatile Multiple-Input, Multiple-Output (MIMO) system. Specifically, by diagonalizing the SSM dynamics, they reparameterized matrix $\overline{\boldsymbol{A}}$ as a diagonal matrix. Simultaneously, $\overline{\boldsymbol{B}}\in\mathbb R^{n\times m}$ and $\overline{\boldsymbol{C}}\in\mathbb R^{n\times m}$ are configured as matrices rather than the vectorized $\overline{\boldsymbol{B}}$ and $\overline{\boldsymbol{C}}$ settings used in S4. However, introducing variable step sizes for different time steps constrains the efficient convolutional computation of the SSM, forcing it to resort to a slower recurrent-based computation. Even with the diagonalized state transition matrix setting, the computational complexity can only be reduced to $O(mL)$, thereby restricted the training efficiency of the SSM.
\paragraph{S6 (Mamba)}The SSM parameters in S4 and S5 are fixed after training, making them data-independent. This somewhat restricts the flexibility of both models. In contrast, S6, as known as Mamba \citep{abs-2312-00752}, overcomes this limitation by introducing data-dependent S4 parameters. It achieves this by employing trainable linear layers to maps the input to each step's $\overline{\boldsymbol{B}}$, $\overline{\boldsymbol{C}}$, and time step $\Delta t$ in S4. Additionally, S6 extended its parameters to be time-variant, transforming from a time-invariant system (as in S4 and S5) to a time-variant one. This enhancement allows S6 to conduct more flexible sequence modeling. However, due to its time-dependent parameterization, S6 cannot efficiently perform SSM computations using convolution, maintaining a computational complexity of $O(mL)$ resulting in slower training compared to S4. 
\section{SSA via State Memory Replay}
In this section, we aim to reveal the problem of NSS in SSM caused by changes in sampling points through ETC theory \citep{HeemelsJT12,Tabuada07} (Section \ref{2.1}). We demonstrate that unstable hidden states lead to errors in SSM (Section \ref{2.2}). Furthermore, through the analysis based on ETC theory, we propose a simple but effective step-size adaptation mechanism, SMR, to enhance the model's SSA capability thus alleviate the NSS problem in SSM with fixed step setting (Section \ref{2.3}). Experimental results indicate that the SMR mechanism can not only enhance the SSA capability of SSM with fixed parameters but can also be extended to other SSM-based models, improving their SSA capabilities (Section \ref{2.4}). 
\subsection{Non-Stable-States Phenomenon}
\label{2.1}
With the help of the ETC theory, we provide a simple example to elucidate the phenomenon of NSS. In this context, ETC theory ensures the system's states remain stable by sampling the input control signal using triggered events. To maintain stability, the selection of sampling points, such as $t_1, t_2, \dots$, must meet specific criteria. Typically, a Lyapunov function $\mathcal L_{V}$ is employed to assess stability \citep{HeemelsJT12}, outside the stable point, it is monotonically decreasing, and the minimum value of $0$ is achieved at the stable point. 
Sampling points that result in a decreasing trend of $\mathcal L_{V}$ are selected to ensure system stability.
Specifically, consider the linear system described in eq.(\ref{eq1}). Assuming the input control signal satisfies the linearity $u(t)=\boldsymbol{T}\boldsymbol{x}(t)$, where $\boldsymbol{T}\in\mathbb R^{m\times n}$, then eq.(\ref{eq1}) becomes:
\begin{equation}
\label{eq7}
\dot{\boldsymbol{x}}(t) ={\boldsymbol{A}} \boldsymbol{x}(t)+{\boldsymbol{B}} \boldsymbol{T}\boldsymbol{x}(t).
\end{equation}
\begin{figure*}[h]

\centering  
\includegraphics[width=1\textwidth]{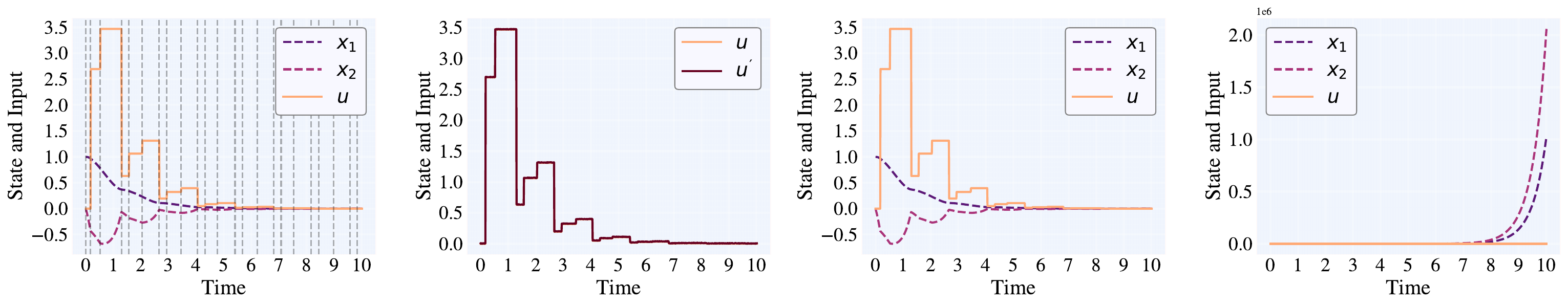}
\caption{An example of the issue of NSS in SSM.}
\label{fig.1}
\end{figure*}
It can be easily verified that $\mathcal L_{V}(t) = \boldsymbol{x}^T \boldsymbol{P} \boldsymbol{x}$ is a Lyapunov function, where symmetric positive definite matrix $\boldsymbol P\in\mathbb R^{n\times n}$ satisfies:
\begin{equation}
\label{eq8}
(\boldsymbol{A}+\boldsymbol{BT})^\top \boldsymbol{P}+ \boldsymbol{P} (\boldsymbol{A}+\boldsymbol{BT})=-\boldsymbol{M},
\end{equation}
to keep $\frac{d\mathcal L_V(t)}{dt}\le 0, \forall \boldsymbol x$, where $\boldsymbol{M}\in \mathbf R^{n\times n}$ is also a symmetric positive definite matrix. Note that the actual sampled input ${u}(t_i)$ is sampled at the sampling points $\{t_i\}_{i\in \mathbb N}$, we denote the sampling error:
\begin{equation}
\label{eq9}
\boldsymbol{e}(t) = \boldsymbol{x}(t_i) - \boldsymbol{x}(t), \quad\forall t\in[t_{i}, t_{i+1}), i\in\mathbb N,    
\end{equation}
then eq.(\ref{eq7}) could be reformulated as:
\begin{equation}
\label{eq10}
\dot{\boldsymbol{x}}(t) = \boldsymbol{A}\boldsymbol{x}(t) + {\boldsymbol{B}} \boldsymbol{T}(\boldsymbol{x}(t)+\boldsymbol{e}(t)).
\end{equation}
Taking the derivative of $\mathcal L_V$, we have:
\begin{equation}
\label{eq11}
\frac{d}{dt}\mathcal L_V(t)= -\boldsymbol{x}(t)^T\boldsymbol M\boldsymbol{x}(t)+2\boldsymbol{x}(t)^\top \boldsymbol{PBT}\boldsymbol{e}(t).   
\end{equation}
Therefore, set $t_0=0$, we have the following triggering condition to ensure system stability:
\begin{equation}
\label{eq12}
\begin{aligned}
  &t_{i+1}=\inf\left\{t\in\mathbb R\mid t>t_i \wedge \kappa  \boldsymbol{x}(t)^\top\boldsymbol M \boldsymbol{x}(t)\right.\\
  &\left.-2\boldsymbol{x}(t)^\top\boldsymbol{PBT}\boldsymbol{e}(t^-)\le 0\right\},
\end{aligned}
\end{equation}
%choosed——>optional
where $\kappa\in (0,1)$ is a optional constant, $e(t^-)$ represents the left-hand limit of error $e$ at point $t$. In other words, new control signals are inputted just before the system becomes unstable. In this way, the sampled input control sequence obtained can ensure exponential stability of the system:
\begin{equation}
\label{eq13}
\mathcal L_V(t)\le \mathcal L_V(0)e^{(\kappa-1)\iota t},
\end{equation}
where $\iota$ is an positive constant. More specifically, we provide an example of a 1-D input where the selected parameters are as follows:
\begin{equation*}
\begin{aligned}
&\boldsymbol A=\begin{bmatrix}
 0 & 1\\
 2 & -3
\end{bmatrix}, \boldsymbol M=\begin{bmatrix}
 0.5 & 0.25\\
 0.25 & 1.5
\end{bmatrix},\\
&\boldsymbol P=\begin{bmatrix}
 1 & 0.25\\
 0.25 & 1
\end{bmatrix}, \boldsymbol B=\begin{bmatrix}
  0\\
  1
\end{bmatrix}, \boldsymbol T=\begin{bmatrix}
 1\\
 -4
\end{bmatrix}, 
\end{aligned}
\end{equation*}

The selected time window is $[0, 10]$, with a time grid width of $0.01$. Subsequently, we conduct simulation experiments on the system, and the results is shown in the leftmost of Fig.\ref{fig.1}, the triggering moment is marked with a gray dashed line. Under the sampled input obtained from ETC, the system's state eventually reaches the stabilization.
\paragraph{NSS: Instability Arising from sampling Grid variation.}  To further substantiate this conclusion, we present an illustrative example. Specifically, we introduce minor perturbations to the sampled data points, strictly constrained within the temporal grid width. The second plot in Fig.\ref{fig.1} illustrates the comparison between the perturbed input and the original input, where the disturbance is almost imperceptible. When utilizing the unaltered sampled data points obtained prior to perturbation as input, the third figure in Fig.\ref{fig.1} visually represents the system's sustained stability. Nevertheless, upon the introduction of perturbated sampled data points into the system, as depicted in the rightmost in Fig.\ref{fig.1}, it becomes apparent that the system's stability cannot be guaranteed, leading to an exponential growth in magnitude reaching $10^6$. This means that when the actual sampling points do not align with the desired sampling grid, it will result in highly unstable states. For SSM models formulated as in eq.(\ref{eq1}) and eq.(\ref{eq2}), encountering such an issue would lead to unavoidable numerical errors (Proposition \ref{prop.1}).

\subsection{Theoretical Understanding of NSS}
\label{2.2}

% Based on the aforementioned considerations and insights, our understandings of S4 are enlightened as follows:
Based on the aforementioned considerations and insights, our understanding of the NSS problem in SSM models is as follows: For SSM models with fixed parameters, the NSS problem may arise when the input does not satisfy stability conditions. Once the sampling error propagates over an extended period along with the hidden states, numerical errors inevitably occur, as affirmed by Proposition \ref{prop.1}.
\begin{figure}[h]
\centering  
\includegraphics[width=0.48\textwidth]{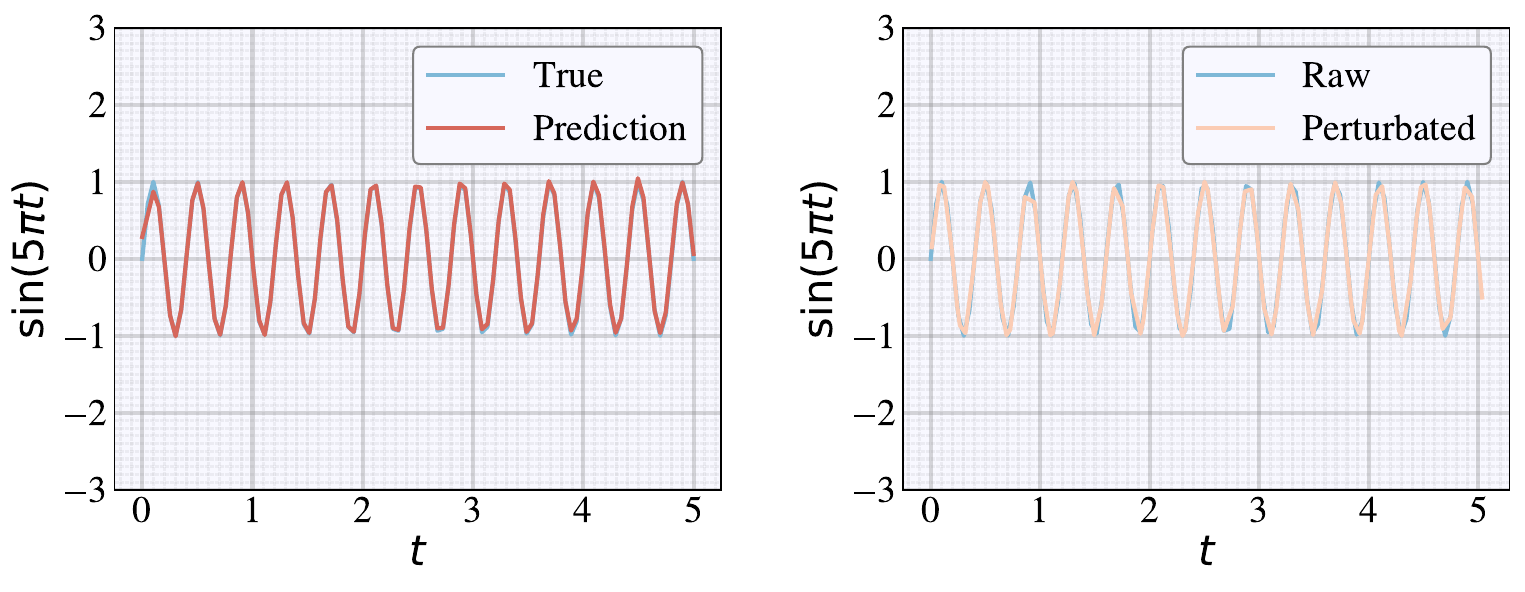}
\caption{An illustrative instance of the NSS issue in S4 is presented here.}
\label{fig.2}
\end{figure}
\newtheorem{proposition}{Proposition}
\begin{proposition}
\label{prop.1} 
Given bounded inputs satisfying $\|{u}\|\le \zeta $, $\|\boldsymbol C\|\le c$ and $\|\boldsymbol B\|<b$, and defining the observation error caused by sampling points as $\boldsymbol{\varepsilon}_i = {u}^\prime_i-{u}_i$, it can be concluded that when $\lim_{t\rightarrow\infty}\|\boldsymbol{x}_t\|>\frac{b\zeta}{1-\left|\lambda_{\max}\right|}$, where $\lambda_{\max}$ represents the largest eigenvalue of matrix $\overline{\boldsymbol{A}}$, the prediction error $\|{y}'_t-{y}_t\|$ will accumulate over time steps.
\end{proposition}
To ascertain the presence of an NSS issue within the SSM model, we devise a simple sequence modeling task. We sample $100$ equidistant points from the function $\sin(5\pi t)$ to serve as input $u$. Then, we employ a single-layer S4 model for fitting, which underwent training for $2000$ epochs, yielding the results displayed in the leftmost in Fig.\ref{fig.2}. 
Following a methodology akin to the previous example, we apply perturbations smaller than the sampling window width to the sampled points $\{t_i\}_{i\in[99]}$. Subsequently, we conduct sampling on the perturbed points $\{t'_i\}_{i\in[99]}$.

This process generates a set of perturbed inputs, denoted as ${u}'$, as illustrated in the second figure in Fig.\ref{fig.2}, where the sampling points underwent slight alterations. Subsequently, we employ the trained S4 model to predict ${u}'$, resulting in a numerical instability, as evident in the third figure in Fig.\ref{fig.2}.

We graphically represent the latent states before and after perturbation in the rightmost figure in Fig.\ref{fig.2}. In both instances, unstable states were observed, and notably, the total magnitude of the state increased following the perturbation. We extend this verification to a $5$-layer S4 model and observe analogous findings. The outcomes are detailed in Appendix \ref{supp.B}.

Therefore, as our analysis reveals, SSMs indeed exhibit the issue of NSS, leading to larger errors when confronted with data exhibiting changes in sampling points. While S5, employing a strategy of assigning different step sizes at each step, can adapt to irregularly sampled data, the fixed SSM parameters during the inference phase still fail to ensure adaptive adjustments to various sampling data, thereby not completely avoiding NSS problems. On the other hand, S6 introduces data-dependent SSM parameterization, ensuring adaptive adjustments during the state transition process. However, this constraint limits S6 from efficiently computing in a convolutional form. In the subsequent analysis, we leverage ETC theory to provide insights and propose a strategy for adaptively adjusting inputs, aiming to address the NSS problem in SSM.

\subsection{State Memory Replay Mechanism}
\label{2.3}
We initiate our investigation by conducting a preliminary analysis rooted in ETC theory to derive insights for formulating adjust strategies.
We examine an input perturbation denoted as $\varepsilon$ at the sampling point, where $u(t+t_{{\varepsilon}})= {u}(t) + \dot {{u}}(t)t_{\varepsilon}+o(t_{\varepsilon})$.
Assuming a tiny perturbation ${\varepsilon}(t)$, we have ${u}'(t) = {u}(t)+{\varepsilon}(t)$.
Hence, the observed state $\boldsymbol{z}(t)$ can be expressed as ${\boldsymbol z}(t) = {\boldsymbol x}(t)-{\boldsymbol e}(t)$, and we also define the discrepancy between the observed state and the actual state as the error $\boldsymbol{e}(t)=\boldsymbol{x}(t)-\boldsymbol{z}(t)$.

\begin{figure}[h]
\centering  
\includegraphics[width=0.45\textwidth]{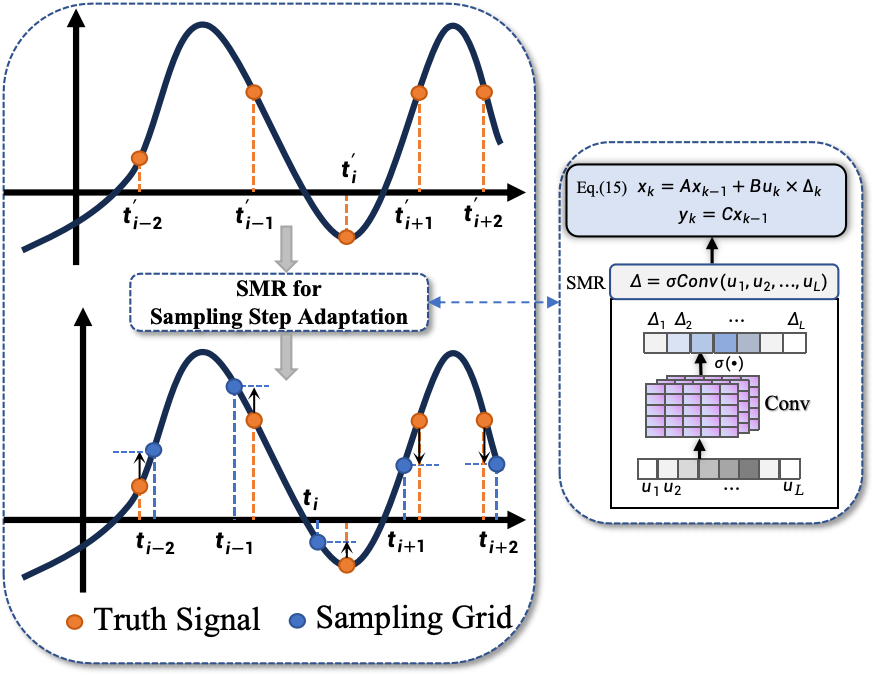}
\caption{Illustration of the proposed SMR Mechanism.}
\label{fig.3}
\vspace{-8pt}
\end{figure}
Drawing inspiration from ETC theory, the Lyapunov function L is utilized as an indicator of observation error stability in the system. A smaller absolute value of $\boldsymbol{e}(t)$ indicates a reduced impact of noise and uncertainty on system performance, as demonstrated in \citep{VallarellaH19}.
Then, we have Theorem \ref{theo.1}.
\newtheorem{theorem}{Theorem}
\begin{theorem}
\label{theo.1}
For the input reply factor $h_\tau (t)=h([t-\tau, t]):[t-\tau,t]\rightarrow \mathbb R$, the adjusted input ${u}_{\text{adj}}(t)=h_\tau (t){u}(t)$, where $\boldsymbol{z}(t)$ is the state value of observer, considering the Lyapunov function $\mathcal{L}_{\boldsymbol{e}}(t)=\boldsymbol{e}^\top(t)\boldsymbol{P}\boldsymbol{e}(t)$, we have:
\begin{equation}
\label{eq14}
\begin{aligned}
&\frac{d\mathcal{L}_{\boldsymbol{e}}(t)}{dt} \le \boldsymbol{e}^\top(t)\left(\boldsymbol{PA}+\boldsymbol{A^\top P}\right)\boldsymbol{e}(t)\\
&  +2\hbar(t)\left(\int_0^t\left\|\boldsymbol{k}(t-l)\right\|\left|\varepsilon(l)\right|dl+\left\|\boldsymbol{B}\right\|\left|\varepsilon(t)\right|\right),    
\end{aligned}
\end{equation}
where $\hbar(t)=\left\|h_{\tau}\right\|_\infty\left\|\boldsymbol{e}(t)\right\|$ and $\boldsymbol{P}$ is a positive definite symmetric
matrix and $\boldsymbol{k}(\cdot):\mathbb R\rightarrow \mathbb R^n$ is a fixed coefficient function determinded by the SSM parameters. 
\end{theorem}
\textbf{Remark 1.}
Theorem \ref{theo.1} suggests that imposing additional constraints on the input controller $h_\tau$ can improve the convergence of the system. In particular, when $h_\tau(\cdot)\equiv 1$ (which corresponding to S4), we have $\left\|h_\tau\right\|_\infty=1$.
The control factor $h_\tau$ is required to incorporate information from the time interval $[t-\tau, t]$.
To accomplish this, a convolution $\text{Conv}_\tau$ with a kernel of length $\tau$, denoted as $\mathcal{K}_{\tau}$, can be utilized.
Moreover, an activation function, denoted as $\sigma$, can be employed to ensure that the condition $\left\|h_\tau\right\|_\infty=\left\|\sigma\circ\text{Conv}_{\tau}\right\|_\infty<1$ is satisfied. This condition contributes to the enhancement of system stability.

To meet this need, considering the analysis in \textbf{Remark 1}, we propose the design of a convolutional learnable variables that incorporates multi input states, enabling adaptive learning and refinement.

\begin{figure}[h]
\centering  
\subfigure{
\includegraphics[width=0.22\textwidth]{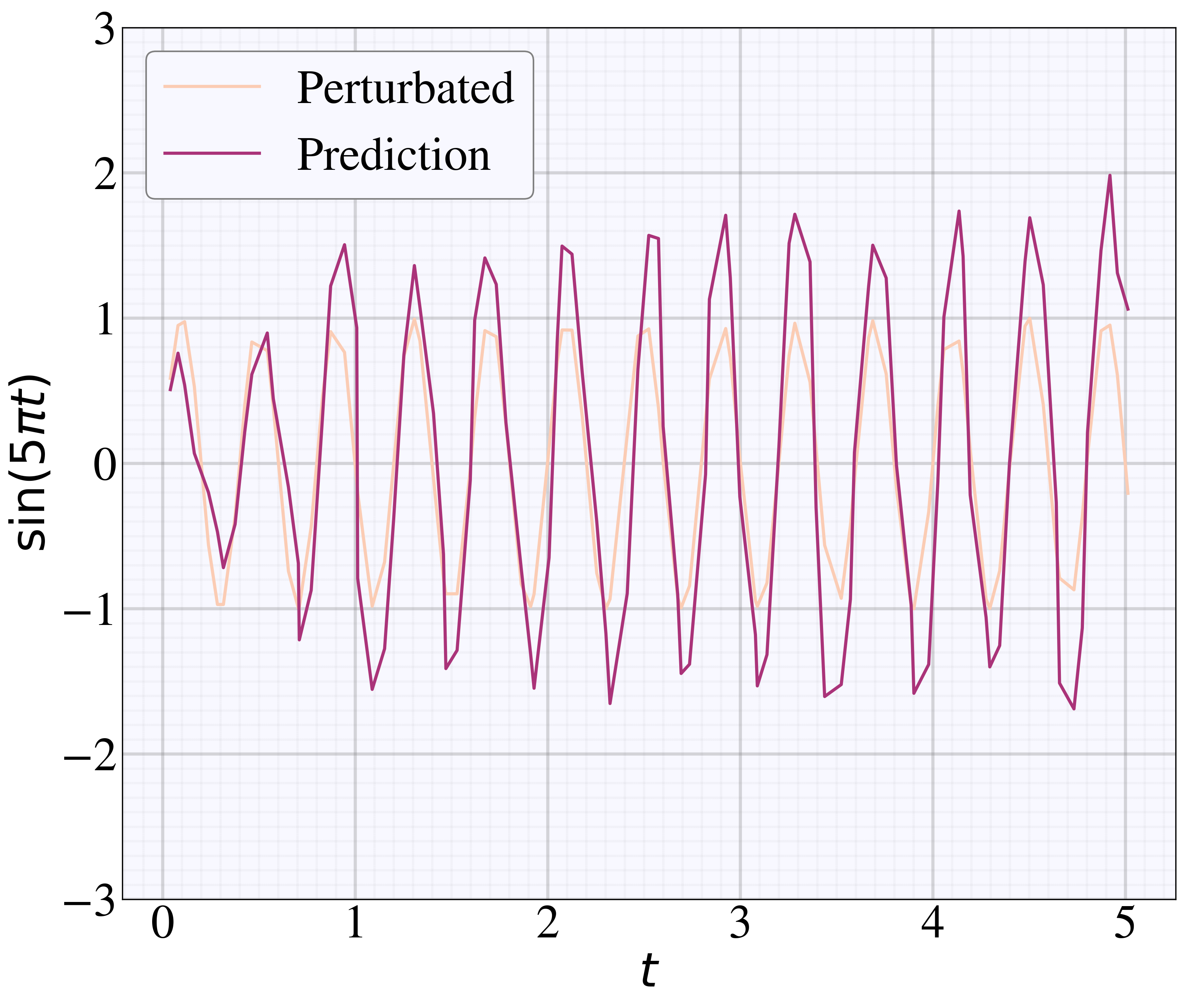}}
\subfigure{
\includegraphics[width=0.22\textwidth]{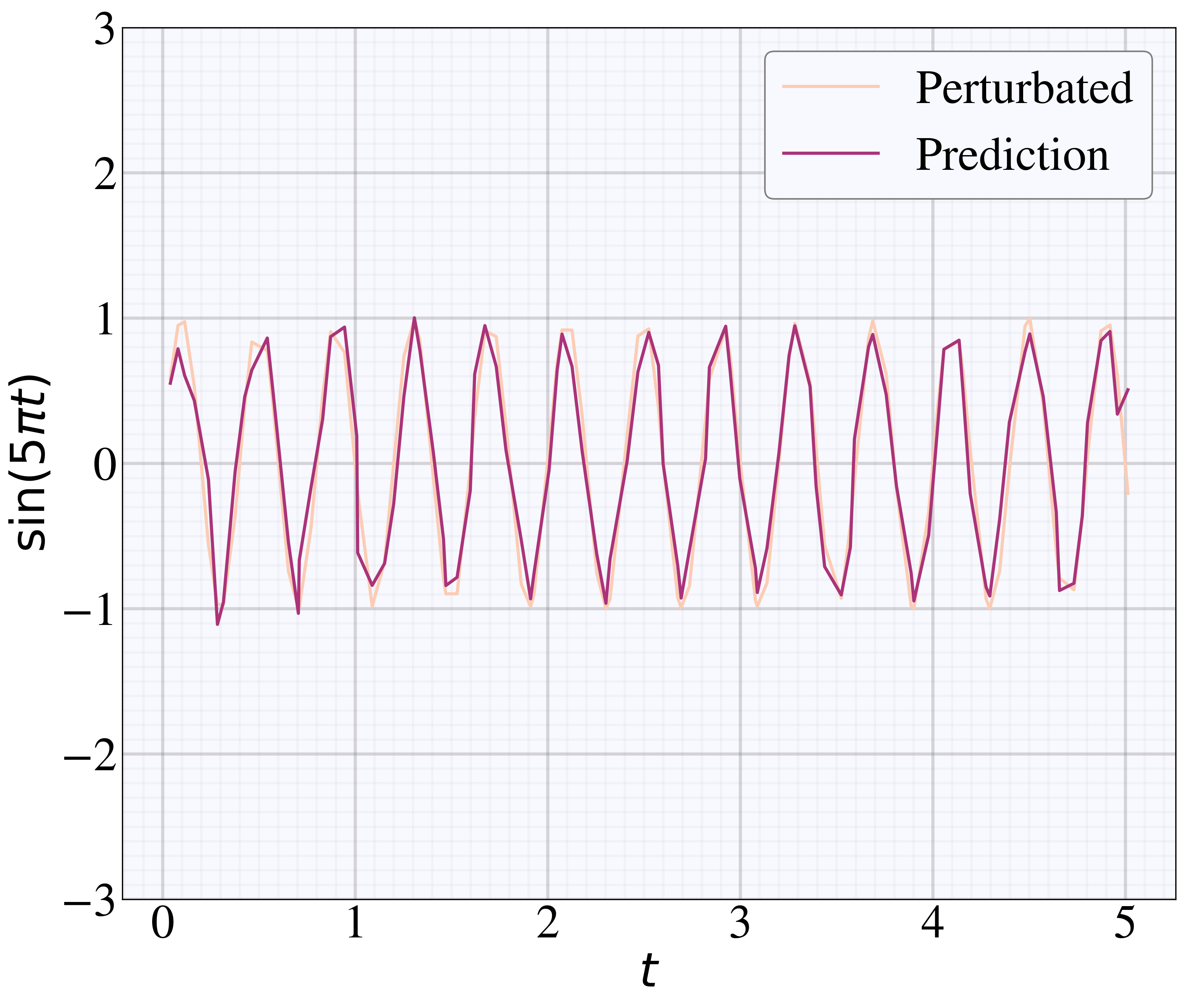}}
\subfigure{
\includegraphics[width=0.22\textwidth]{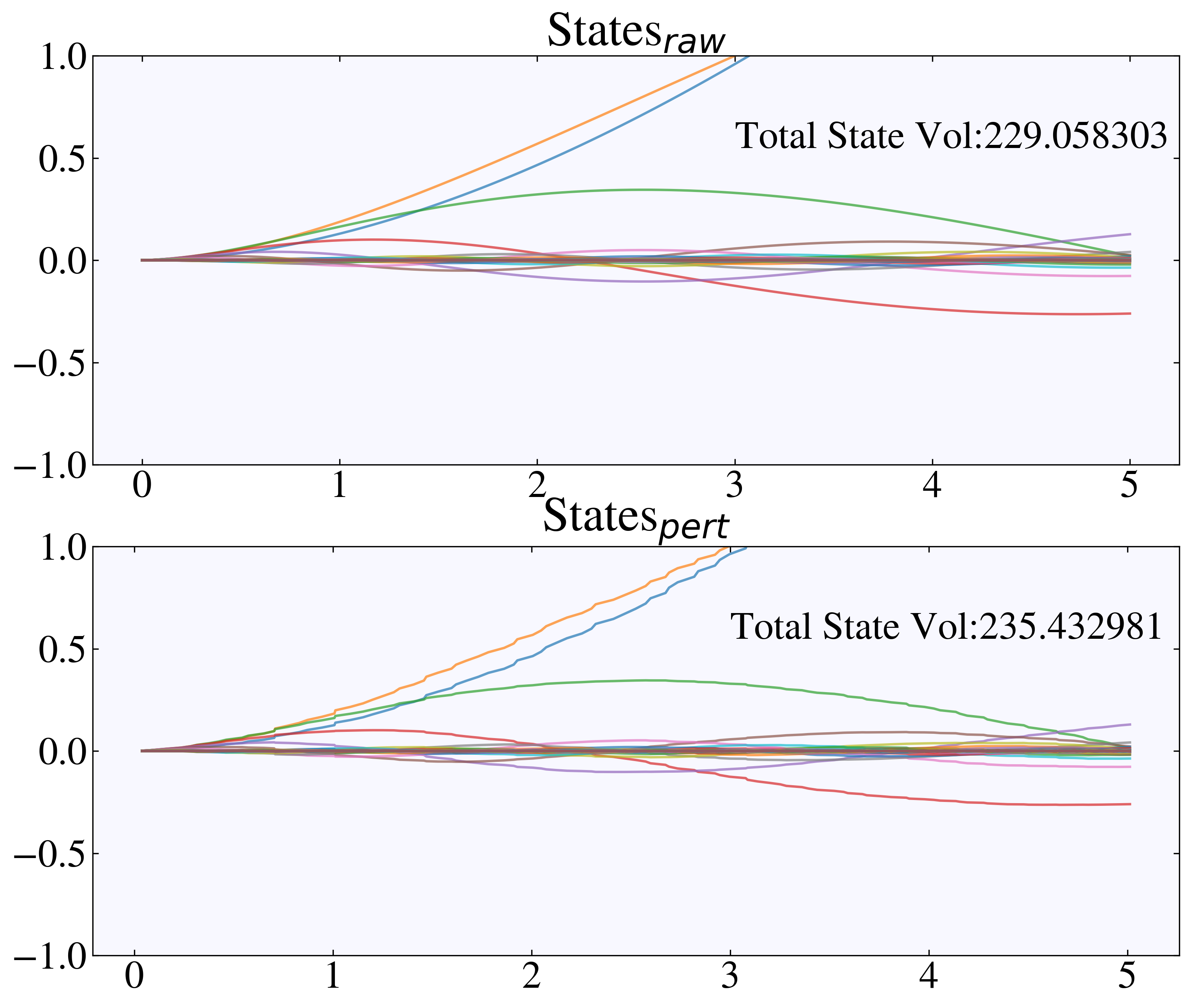}}
\subfigure{
\includegraphics[width=0.22\textwidth]{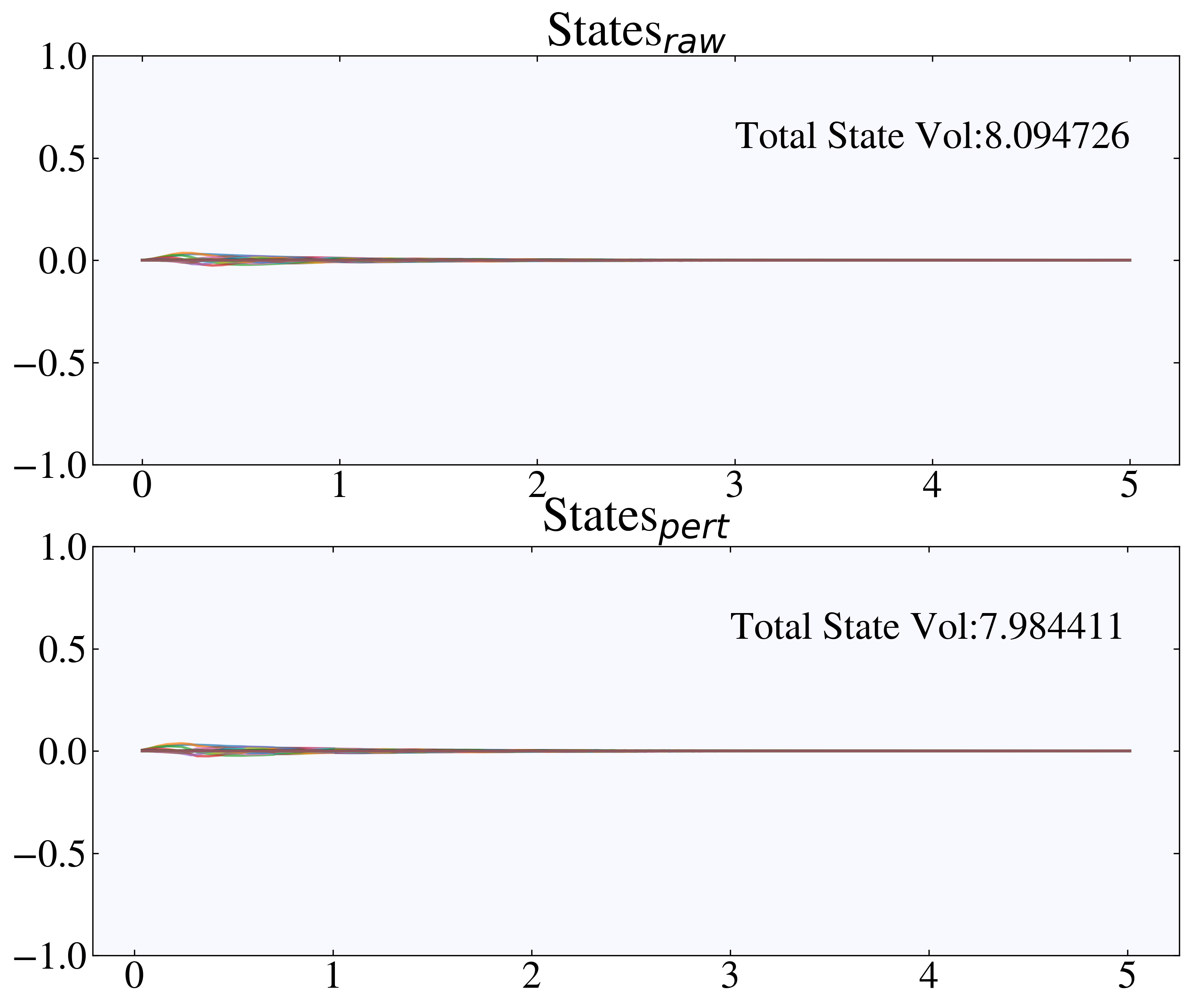}}
\caption{Comparative results of S4 incorporated with SMR (S4+SMR) on the aforementioned examples. The pair of figures displays the prediction outcomes of S4+SMR for the perturbed input $u'$ (left) and the latent states when provided with inputs $u$ and $u'$ (right).}
\label{fig.4}
\vspace{-5pt}
\end{figure}

\begin{figure*}[t]
\centering  
\includegraphics[width=0.8\textwidth]{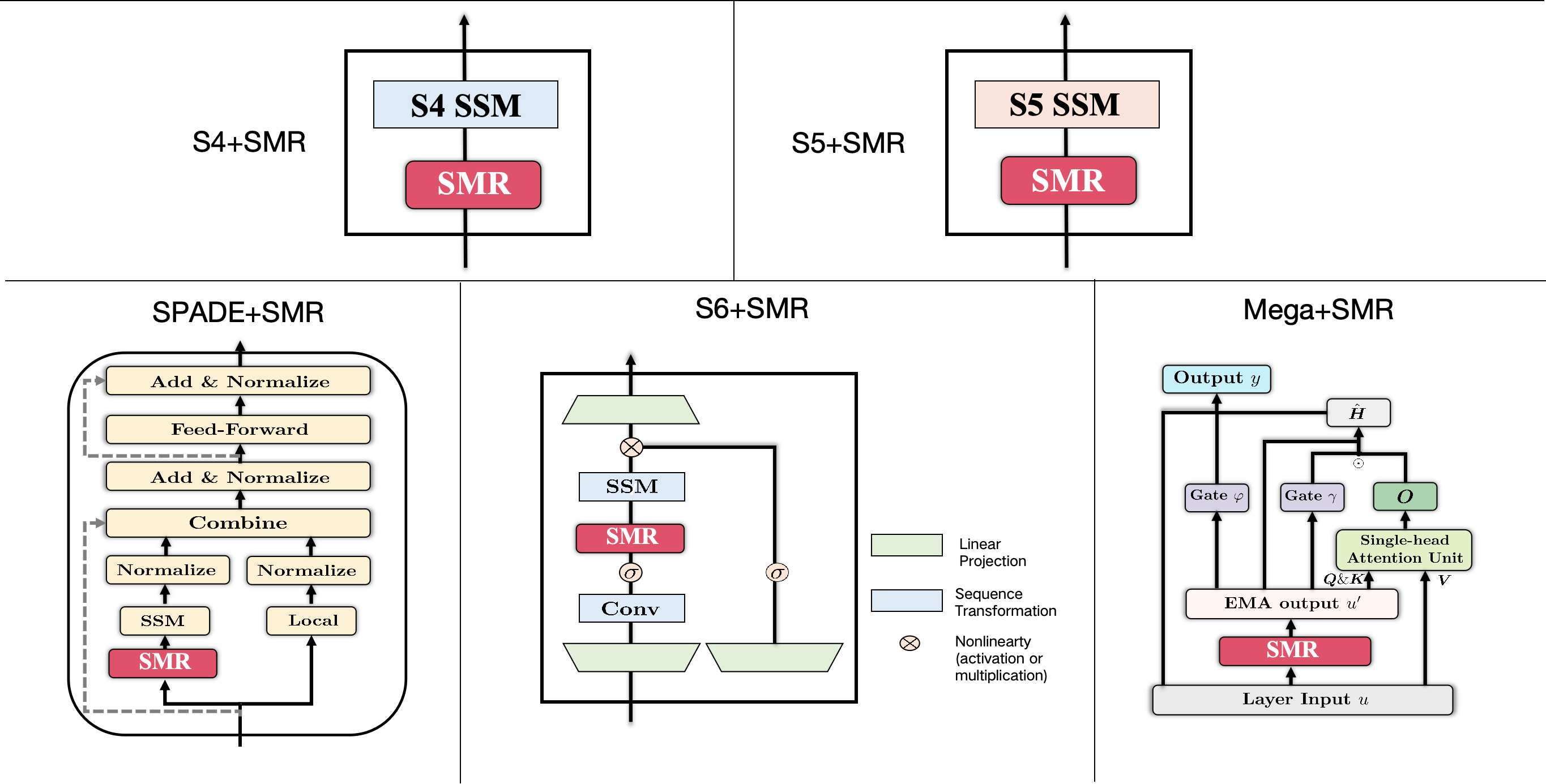}
\caption{Schematic diagram of various SSMs after incorporating SMR.}
\label{fig.10}
\vspace{-10pt}
\end{figure*}

Building upon Theorem \ref{theo.1}, we understand the importance of having learnable variables that can incorporate multi input states to control how sampling information behaves, allowing for automatic adjustments. 
To fulfill this requirement, considering the analysis in \textbf{Remark 1}, we propose the SMR mechanism aimed at addressing the NSS problem caused by variations in sampling points. The SMR mechanism incorporates learnable memories to enhance the SSM model with multiple memory steps, through a convolutional learnable variables that incorporates multi input steps, enabling adaptive learning and refinement, as depicted in Fig.\ref{fig.3}. Formally,our proposed SMR mechanism can be formulated as:
\begin{equation}
\label{eq15}
\begin{aligned}
&x_{k}=\overline{\boldsymbol{A}} \boldsymbol{x}_{k-1}+\\
&\overline{\boldsymbol{B}} u_{k}\sigma_{\text{Sig}}(\mathcal{K}_\tau \ast (\underbrace{u_{1},\dots, u_{1}}_{\tau},\dots,u_T))_k,
\end{aligned}
% \vspace{-5pt}
\end{equation}
% \vspace{-5pt}
where $\tau$ represents the convolutional kernel length, and $\sigma_{\text{Sig}}(\cdot)$ refers to the Sigmoid function.
In particular, integrating SMR into S4 ensures the efficient computation of SSM through convolutional operations. Simultaneously, it introduces enhanced flexibility to the SSM, enabling it to adapt to diverse sampling intervals and changing sample points. To validate the efficacy of SMR in mitigating NSS issues in SSMs, we conduct training and testing by incorporating SMR into the 1-layer S4 model, following the previously mentioned experimental configurations. The results are presented in Fig.\ref{fig.4}.
The model's fitting results on $u'$ is displayed in the left of Fig.\ref{fig.4}, demonstrating the successful mitigation of unstable numerical outputs and a substantial reduction in prediction errors. The results illustrate in the second figure of Fig.\ref{fig.4} clearly indicate that the latent states of S4+SMR have achieved stability, characterized by a significantly reduced total volume of the absolute state values, shrinking from $2\times 10^2$ as shown in Fig.\ref{fig.2} to $7.98$.
This implies that the integration of SMR significantly addresses the NSS issues in S4. Furthermore, experiments conducted on a 5-layer S4+SMR architecture also showed alleviation of NSS issues and improved predictive accuracy on perturbed data, the detailed results are presented in Appendix \ref{supp.B}.
% \vspace{-8pt}
By incorporating the SMR (as its code shown in the code in List \ref{lst:smr_class}) into the SSMs at the positions indicated in Fig. \ref{fig.10}, it is easily to integrate the SMR into a variety of SSMs.

\subsection{Empirical Validation of SMR for SSA}
\label{2.4}
To further investigate the impact of the SMR mechanism on enhancing the SSM model's SSA capability, we utilize a Pendulum dataset \citep{SchirmerELR22,SmithWL23} characterized by irregularly sampled points and varying sampling intervals, to construct a regression task. The dataset comprises sequences of pendulum images with a length of $L=50$ as input. Each image, sized $24\times 24$, is sampled at non-uniform time intervals ranging from $T=0$ to $T=100$. Notably, the sampling points for each data instance exhibit variability. Some images in the sequence are intentionally corrupted by random noise, introducing "occlusion" and resulting in more irregular sampling trajectories. The prediction target $y_{tar}\in\mathbb R^{50\times 2}$ is the sine and cosine values corresponding to the position of the pendulum in each image of the input sequence $u\in\mathbb R^{50\times 576}$. Examples of this dataset can be found in Appendix \ref{Appendix C}.

To prevent model overfitting at each time point due to an excessive amount of constructed training data, ensuring that only models with strong generalization capabilities for changing sample points can effectively handle the task, we opt for a more challenging setup compared to the setting in \citep{SchirmerELR22} with $2000$ training data and $1000$ testing data points. Specifically, we allocate $500$ training data sequences and $200$ testing data sequences to make the task more challenging. In this task, we conduct comparative experiments with S4, both with and without the SMR mechanism. Additionally, to explore the generalization of our SMR mechanism to a broader range of SSM-based models, we include the more flexible SSM models, S5 and S6, in the comparison. Furthermore, we select two models that combine Attention with convolution-based SSM, Mega \cite{MaZKHGNMZ23} and SPADE \cite{abs-2212-08136}, known for their competitive language modeling and long sequence modeling capabilities. We integrate SMR before Mega's EMA operation and before SPADE's S4 module to investigate the impact of SMR on various structures of SSMs. 

For the given model $\mathcal{M}$, we choose the Mean Squared Error (MSE) computed on the test set $\mathcal{V}$, i.e., $\frac{1}{|\mathcal{V}|}\sum_{\{u,y_{tar}\}\in \mathcal{V}}{(\mathcal{M}(u) - y_{tar})^2}$, as the evaluation criterion. We report the best result obtained throughout $100$ training epochs in Tab.\ref{tab_irr}. The integration of SMR brings about a significant improvement in S4's SSA capability. Notably, this enhancement is not exclusive to S4, even S5 shows a considerable performance boost upon incorporating SMR. To be more specific, the test MSE decreases by $8.31$ for S5, indicating that SMR significantly improves S5's capability to handle variations in sampling points. Additionally, S6, SPADE, and Mega all demonstrate a decrease in Test MSE after integrating SMR. This suggests that our proposed SMR not only assists convolution-based SSMs in enhancing its SSA capability but also generalizes to recurrence-based SSMs, offering widespread improvements.

\lstset{
  language=Python,
  basicstyle=\small\ttfamily, 
  keywordstyle=\color{blue!70}\bfseries, 
  commentstyle=\color{green!70},
  stringstyle=\color{red!70}, 
  showstringspaces=false, 
  frame=single, 
  backgroundcolor=\color{lightgray!20}, 
  rulesepcolor=\color{gray!50},
  tabsize=4, 
  breaklines=true, 
  showspaces=false, 
  captionpos=b, 
}

\begin{figure*} 
\centering
\begin{lstlisting}[language=Python, caption={The code of SMR}, label=lst:smr_class]
class SMR(nn.Module):
    def __init__(self, in_features, out_features, kernel_size, linear = False):
        super(SMR, self).__init__()
        self.conv = nn.Conv1d(in_features, out_features, kernel_size, stride=1)
        self.use_linear = linear
        if linear:
            self.linear = nn.Linear(in_features, out_features)
        self.pad = (kernel_size - 1, 0)
    def forward(self, x):
        # Input shape: (B, H, L) 
        # Output shape: (B, H, L)
        if self.use_linear:
            factor = self.linear(self.conv(F.pad(x, self.pad, mode='constant', value=0.0)).transpose(1, 2)).transpose(1, 2)
        else:
            factor = self.conv(F.pad(x, self.pad, mode='constant', value=0.0))
        return torch.sigmoid(factor) * x 
\end{lstlisting}
\vspace{-10pt}
\end{figure*} 

\begin{table}[h]
\renewcommand{\arraystretch}{0.6}
  \centering
  \small
  \caption{The test MSE on the pendulum dataset, where "\textbf{w/ SMR}" and "w/o SMR" respectively indicate the cases with and without the incorporation of the SMR mechanism. "Mode" represents the computation mode of the SSM.}
   \begin{tabular}{c|c|c|c}
        \hline
        \multirow{2}{*}{Mode} & \multirow{2}{*}{Model}& \multicolumn{2}{c}{Test MSE} \\
        \cline{3-4}
        & & \multicolumn{1}{c|}{w/o SMR} & \multicolumn{1}{c}{\textbf{w/ SMR}} \\
    \midrule
    \multirow{3}{*}{Convolution} & S4 & 10.99 & \textbf{2.14} \\
     & Mega & 1.72 & \textbf{1.61} \\
     & SPADE & 2.58 & \textbf{2.17} \\
    \midrule
    \multirow{2}{*}{Recurrence} & S5 & 10.40 & \textbf{2.09} \\
     & S6 & 5.17 & \textbf{4.46} \\
    \bottomrule
    \end{tabular}%
  \label{tab_irr}%
\end{table}

\begin{table}[h]
\renewcommand{\arraystretch}{0.6}
\setlength{\tabcolsep}{0.95pt}
  \centering
  \small
  \caption{Perplexity (PPL) on Wikitext-103. The results on the left and right of "/" correspond to w/o SMR and \textbf{w/ SMR}, respectively. "Mode" represents the computation mode of the SSM.}
    \begin{tabular}{c|c|cc}
    \toprule
    Mode & Model & PPL(val) & PPL(test) \\
    \midrule
    - &Trans  & 24.42 & 24.81 \\
    - &LS & 23.71  &  24.13 \\
    \hline
    \multirow{3}{*}{Convolution} &S4 & 39.32/\textbf{36.48} & 40.02/\textbf{38.16} \\
    &Mega & 26.30/\textbf{25.28} & 26.75/\textbf{25.67} \\
    &SPADE & 24.18/\textbf{23.68} & 24.55/\textbf{23.99} \\
    \hline
    \multirow{2}{*}{Recurrence} &S5 & 33.52/\textbf{33.29} & 35.09/\textbf{34.72} \\
    &S6 & 23.97/\textbf{23.85} & 24.95/\textbf{24.78} \\
    \bottomrule
    \end{tabular}%
  \label{tab2}%
  \vspace{-8pt}
\end{table}

\begin{table*}[t]
 \small
  \renewcommand{\arraystretch}{0.6}
   \centering
   \caption{Experimental results on the LRA Benchmark. The results on the left and right of "/" correspond to w/o SMR and \textbf{w/ SMR}, respectively. "Mode" represents the computation mode of the SSM.} 
     \begin{tabular*}{1\textwidth}{@{\extracolsep{\fill}}c|c|cccccc}
    \hline
    \multicolumn{1}{c|}{Mode}&\multicolumn{1}{c|}{Model} & \multicolumn{1}{c}{Text} & \multicolumn{1}{c}{ ListOps} & \multicolumn{1}{c}{Retrieval} & \multicolumn{1}{c}{Image} & \multicolumn{1}{c}{Pathfinder}  & \multicolumn{1}{c}{AVG} \\
    \hline
    - &Transformer& 61.95 & 38.37 & 80.69 & 65.26 & 40.57 & 57.37 \\
    - &LS & 66.62 & 40.30 & 81.68 & 69.98 &  47.60 & 61.24 \\
    \hline
    \hline
    \multirow{3}{*}{Convolution}& S4 &  86.47/\textbf{89.09} & 57.06/\textbf{59.01} & 86.74/\textbf{89.28} &  87.20/\textbf{88.97} & 85.99/\textbf{89.01}  &  80.69/\textbf{83.07} \\
    & Mega & 89.97/\textbf{90.36} & 57.67/\textbf{59.45} & 90.17/\textbf{90.64} & 86.82/\textbf{88.21} & 93.40/\textbf{93.78} & 83.61/\textbf{84.49} \\
    & SPADE & 86.29/\textbf{87.06} & 58.75/\textbf{59.52} & 88.62/\textbf{89.01} & 88.05/\textbf{89.29} & 92.77/\textbf{93.34} & 82.90/\textbf{83.64} \\
    \hline
    \hline
    \multirow{2}{*}{Recurrence} & S5 & 84.20/\textbf{87.08} & 58.25/\textbf{59.08} & 87.99/\textbf{89.37} & 87.51/\textbf{89.31} & 87.42/\textbf{88.05} & 81.07/\textbf{82.58} \\
    & S6 & 83.52/\textbf{84.14} & 55.62/\textbf{56.15} & 83.28/\textbf{83.66} & 82.96/\textbf{83.23} & 85.54/\textbf{85.80} & 78.18/\textbf{78.60} \\
    
    \hline
\end{tabular*}%
   \label{tab4}%
    
 \end{table*}%
\section{Experiments}
\label{4}
As stated previously, the integration of SMR further enhances SSM's SSA capability, thereby providing increased flexibility in sequence modeling capabilities. To further assess the improvement in sequential modeling capacity brought about by SMR for SSM-based models, we have chosen two more practical sequence modeling tasks: autoregressive language modeling and long-term dependency modeling. Our experimental setup follows that outlined in Section \ref{2.4}. For S4 \citep{GuGR22}, S5 \citep{SmithWL23}, S6 \citep{abs-2312-00752}, SPADE \citep{abs-2212-08136} and Mega \citep{MaZKHGNMZ23}, we conducted ablation experiments with and without SMR inclusion to evaluate the generalizability benefits that the SMR mechanism confers upon SSM-based models in these sequence modeling tasks. To better illustrate the significance of the benefits brought by SMR, we introduced the comparative results on the respective tasks the Vanilla Transformer \citep{vaswani2017attention} and the state-of-the-art (on WikiText-103) Transformer-based model, Transformer-LS (LS) \citep{ZhuPXSGAC21}. All experiments were conducted on four Tesla A800 GPUs.

\subsection{Autoregressive language modeling}
\label{4.1}

To evaluate the ability of autoregressive language modeling, we conducted experiments on the WikiText-103 dataset \citep{MerityX0S17}. This dataset comprises 103 million word-level tokens extracted from Wikipedia articles.
In accordance with \citep{QinHSHLLDKZ23}, all models were trained on the WikiText-103 dataset for $50,000$ steps, using a learning rate of $5e-4$.
The sequence length is set to $512$, and weight decay is set to $0.1$ for all models. Consistent with the configuration detailed in \citep{Chen21}, all models were uniformly set up with six layers and a hidden dimension of $512$. The performance of autoregressive language modeling is assessed by reporting perplexity (PPL) scores on both the validation and test sets. For more detailed information regarding the experiments, please refer to Appendix \ref{Appendix C}.

Tab.\ref{tab2} showcases consistent improvements in both validation and test perplexity (PPL) for all SSM-based models subjected to the experiments after incorporating the SMR mechanism. While S4, due to its fixed parameters and constant time-step settings, faces limitations in language tasks, integrating SMR yields a significant reduction of 2.84 and 1.86 in validation and test PPL, respectively. Notably, SMR incorporation in SPADE leads to a further 0.56 decrease in test PPL, even surpassing the performance of Transformer-LS. These findings solidify that the SMR mechanism enhances the flexibility of SSM models, ultimately contributing to advancements in the autoregressive language modeling capabilities of SSM-based architectures.

\subsection{Long-range dependency modeling}
\label{4.3}

To further assess the impact of SMR on long sequence modeling, we conducted experiments on five Long Range Arena (LRA) benchmark tasks: ListOps \citep{NangiaB18}, Byte-level Text Classification \citep{MaasDPHNP11}, Byte-level Document Retrieval \citep{RadevMQA13}, Sequence CIFAR-10 \citep{2009Learning}, and Pathfinder \citep{LinsleyKVWS18}. All models used consistent block and hidden dimension settings for each task. Detailed configurations in Appendix \ref{Appendix D}. Results in Tab.\ref{tab4} demonstrate that SMR integration consistently improves the performance of various SSM-based models. Notably, SMR achieves an average performance gain of 2.38 and 1.51 on tasks S4 and S5, respectively. Furthermore, SMR contributes to performance improvements in models S6, Mega, and SPADE. These findings suggest that SMR universally enhances the long sequence modeling capabilities of SSM-based models.

\subsection{The Impact of SMR on Training Speed}
To propose a strategy that improves the flexibility of SSM without impacting their training efficiency, we investigated whether integrating the SMR mechanism could enhance sequence modeling capabilities while maintaining training speed. Therefore, we conducted experiments on the Wikitext-103 dataset, comparing the relative training speed ratios of various models with and without the SMR mechanism. Due to the fact that our implementation of S4 and S5 were solely based on torch without utilizing the acceleration provided by related CUDA extension, we included a version of S6 implemented purely with torch as a baseline (1.0$\times$ speed) for a more direct speed comparison between models. Experimental results, presented in Tab.\ref{tab:speed}, demonstrate that SMR incorporation does not significantly decrease SSM training speed and preserves the relative speed relationships among different SSM-based models. This suggests SMR serves as an effective way to enhance the sequence modeling capabilities of SSM without compromising its training efficiency.

\begin{table}[h]
  \centering
  \renewcommand{\arraystretch}{0.6}
  \small
  \caption{Comparison of training speeds on Wikitext-103. We use the S6 implemented purely in torch incorporated as the baseline (1.0$\times$) and report the relative training speed ratios with respect to this value. "Mode" represents the computation mode of the SSM.}
   \begin{tabular}{c|c|c|c}
        \hline
        \multirow{2}{*}{Mode} & \multirow{2}{*}{Model} & \multicolumn{2}{c}{Relative Speed} \\
        \cline{3-4}
       & & \multicolumn{1}{c|}{w/o SMR} & \multicolumn{1}{c}{\textbf{w/ SMR}} \\
    \midrule
    \multirow{3}{*}{Convolution} &S4 & 8.72$\times$ & \textbf{8.43}$\times$ \\
    &Mega  & 6.48$\times$ & \textbf{6.31}$\times$ \\
    &SPADE  & 7.29$\times$ & \textbf{6.92}$\times$ \\
    \hline
    \multirow{3}{*}{Recurrence}& S6 (in torch)\tablefootnote{\url{https://github.com/alxndrTL/mamba.py.git}} & 1.0$\times$ & - \\
    &S5  & 6.18$\times$ & \textbf{5.87}$\times$ \\
    &S6  & 2.99$\times$ & \textbf{2.49}$\times$ \\
    \bottomrule
    \end{tabular}%
    \vspace{-8pt}
  \label{tab:speed}%
\end{table}
\section{Conclusion}
In this paper, we investigated the NSS issue in SSMs for long sequence modeling, we found that when input data deviates from the model's sampling requirements, it leads to error accumulation and hidden state divergence. Our analysis further revealed that early memory adjustments in the input sequence can achieve adaptive sampling, effectively solving the NSS problem. Inspired by this, we proposed a simple yet efficient plug-and-play mechanism, SMR. Theoretical analysis and experiments demonstrated that SMR effectively alleviates NSS, enhancing the generalization ability of SSMs to diverse sampling points and leading to superior sequence modeling performance.
We evaluated SMR on various SSM-based models, including the convolution-based and recurrence-based SSMs, applying it to both autoregressive language modeling (on Wikitext-103) and the LRA benchmark. The results demonstrate that SMR significantly improves the performance of SSM-based models on these tasks, solidifying its effectiveness and broad applicability.

\section{Limitations}
This study investigates the NSS issue of SSMs for long sequence modeling from a novel theoretical perspective of ETC theory. We first conduct preliminary experimental analysis and theoretical verification to validate the existence of NSS. Inspired by the analysis, we design a simple yet effective SMR mechanism and verify its effectiveness on datasets with different sampling resolutions. Furthermore, experiments demonstrate significant improvements on convolution-based SSMs S4, Mega and SPADE, as well as recurrence-based SSMs S5 and S6 on benchmarks such as wikitext and LRA.

However, the current study is preliminary. In the future, we can extend this technology to interactive learning frameworks \cite{abs-2403-02628}, explore continual SSM frameworks \cite{qi2024contrastive}, and design more robust and secure models \cite{abs-2402-16397,GaoQLGLXZ23,QiGLWZ24}, applying them to scenarios such as knowledge discovery \cite{qi2023large}.

In conclusion, our research points out the NSS issue in SSMs and demonstrates that incorporating this factor into new long sequence model architectures is a promising direction that requires extensive exploration. We believe that these new findings can better promote the optimization and upgrading of SSM-based architectures.

\section{Ethics Statement}
The purpose of this paper is technical research, and the tasks, models, and datasets involved do not raise any ethical or moral concerns.

\section{acknowledgement}
This work was supported in part by the National Science and Technology Major Project (No. 20232D0121403). We extend our gratitude to the anonymous reviewers for their insightful feedback, which has greatly contributed to the improvement of this paper.

\bibliography{main}

\begin{thebibliography}{34}
\expandafter\ifx\csname natexlab\endcsname\relax\def\natexlab#1{#1}\fi

\bibitem[{Beltagy et~al.(2020)Beltagy, Peters, and Cohan}]{beltagy2020longformer}
Iz~Beltagy, Matthew~E Peters, and Arman Cohan. 2020.
\newblock Longformer: The long-document transformer.
\newblock \emph{arXiv preprint arXiv:2004.05150}.

\bibitem[{Brown et~al.(2020)Brown, Mann, Ryder, Subbiah, Kaplan, Dhariwal, Neelakantan, Shyam, Sastry, Askell, Agarwal, Herbert{-}Voss, Krueger, Henighan, Child, Ramesh, Ziegler, Wu, Winter, Hesse, Chen, Sigler, Litwin, Gray, Chess, Clark, Berner, McCandlish, Radford, Sutskever, and Amodei}]{BrownMRSKDNSSAA20}
Tom~B. Brown, Benjamin Mann, Nick Ryder, Melanie Subbiah, Jared Kaplan, Prafulla Dhariwal, Arvind Neelakantan, Pranav Shyam, Girish Sastry, Amanda Askell, Sandhini Agarwal, Ariel Herbert{-}Voss, Gretchen Krueger, Tom Henighan, Rewon Child, Aditya Ramesh, Daniel~M. Ziegler, Jeffrey Wu, Clemens Winter, Christopher Hesse, Mark Chen, Eric Sigler, Mateusz Litwin, Scott Gray, Benjamin Chess, Jack Clark, Christopher Berner, Sam McCandlish, Alec Radford, Ilya Sutskever, and Dario Amodei. 2020.
\newblock \href {https://proceedings.neurips.cc/paper/2020/hash/1457c0d6bfcb4967418bfb8ac142f64a-Abstract.html} {Language models are few-shot learners}.
\newblock In \emph{Advances in Neural Information Processing Systems 33: Annual Conference on Neural Information Processing Systems 2020, NeurIPS 2020, December 6-12, 2020, virtual}.

\bibitem[{Chen(2021)}]{Chen21}
Peng Chen. 2021.
\newblock \href {https://doi.org/10.18653/V1/2021.EMNLP-MAIN.828} {Permuteformer: Efficient relative position encoding for long sequences}.
\newblock In \emph{Proceedings of the 2021 Conference on Empirical Methods in Natural Language Processing, {EMNLP} 2021, Virtual Event / Punta Cana, Dominican Republic, 7-11 November, 2021}, pages 10606--10618. Association for Computational Linguistics.

\bibitem[{Choromanski et~al.(2020)Choromanski, Likhosherstov, Dohan, Song, Gane, Sarlos, Hawkins, Davis, Mohiuddin, Kaiser et~al.}]{choromanski2020rethinking}
Krzysztof~Marcin Choromanski, Valerii Likhosherstov, David Dohan, Xingyou Song, Andreea Gane, Tamas Sarlos, Peter Hawkins, Jared~Quincy Davis, Afroz Mohiuddin, Lukasz Kaiser, et~al. 2020.
\newblock Rethinking attention with performers.
\newblock In \emph{International Conference on Learning Representations}.

\bibitem[{Gao et~al.(2023)Gao, Qi, Li, Guo, Li, Xing, and Zhang}]{GaoQLGLXZ23}
Junqi Gao, Biqing Qi, Yao Li, Zhichang Guo, Dong Li, Yuming Xing, and Dazhi Zhang. 2023.
\newblock \href {http://papers.nips.cc/paper\_files/paper/2023/hash/028fcbcf85435d39a40c4d61b42c99a4-Abstract-Conference.html} {Perturbation towards easy samples improves targeted adversarial transferability}.
\newblock In \emph{Advances in Neural Information Processing Systems 36: Annual Conference on Neural Information Processing Systems 2023, NeurIPS 2023, New Orleans, LA, USA, December 10 - 16, 2023}.

\bibitem[{Gu and Dao(2023)}]{abs-2312-00752}
Albert Gu and Tri Dao. 2023.
\newblock \href {https://doi.org/10.48550/ARXIV.2312.00752} {Mamba: Linear-time sequence modeling with selective state spaces}.
\newblock \emph{CoRR}, abs/2312.00752.

\bibitem[{Gu et~al.(2021{\natexlab{a}})Gu, Goel, and Re}]{gu2021efficiently}
Albert Gu, Karan Goel, and Christopher Re. 2021{\natexlab{a}}.
\newblock Efficiently modeling long sequences with structured state spaces.
\newblock In \emph{International Conference on Learning Representations}.

\bibitem[{Gu et~al.(2022)Gu, Goel, and R{\'{e}}}]{GuGR22}
Albert Gu, Karan Goel, and Christopher R{\'{e}}. 2022.
\newblock \href {https://openreview.net/forum?id=uYLFoz1vlAC} {Efficiently modeling long sequences with structured state spaces}.
\newblock In \emph{The Tenth International Conference on Learning Representations, {ICLR} 2022, Virtual Event, April 25-29, 2022}. OpenReview.net.

\bibitem[{Gu et~al.(2021{\natexlab{b}})Gu, Johnson, Goel, Saab, Dao, Rudra, and R{\'{e}}}]{GuJGSDRR21}
Albert Gu, Isys Johnson, Karan Goel, Khaled Saab, Tri Dao, Atri Rudra, and Christopher R{\'{e}}. 2021{\natexlab{b}}.
\newblock \href {https://proceedings.neurips.cc/paper/2021/hash/05546b0e38ab9175cd905eebcc6ebb76-Abstract.html} {Combining recurrent, convolutional, and continuous-time models with linear state space layers}.
\newblock In \emph{Advances in Neural Information Processing Systems 34: Annual Conference on Neural Information Processing Systems 2021, NeurIPS 2021, December 6-14, 2021, virtual}, pages 572--585.

\bibitem[{Gupta et~al.(2022)Gupta, Gu, and Berant}]{gupta2022diagonal}
Ankit Gupta, Albert Gu, and Jonathan Berant. 2022.
\newblock Diagonal state spaces are as effective as structured state spaces.
\newblock \emph{Advances in Neural Information Processing Systems}, 35:22982--22994.

\bibitem[{Heemels et~al.(2012)Heemels, Johansson, and Tabuada}]{HeemelsJT12}
W.~P. M.~H. Heemels, Karl~Henrik Johansson, and Paulo Tabuada. 2012.
\newblock \href {https://doi.org/10.1109/CDC.2012.6425820} {An introduction to event-triggered and self-triggered control}.
\newblock In \emph{Proceedings of the 51th {IEEE} Conference on Decision and Control, {CDC} 2012, December 10-13, 2012, Maui, HI, {USA}}, pages 3270--3285. {IEEE}.

\bibitem[{Krizhevsky and Hinton(2009)}]{2009Learning}
A.~Krizhevsky and G.~Hinton. 2009.
\newblock Learning multiple layers of features from tiny images.
\newblock \emph{Handbook of Systemic Autoimmune Diseases}, 1(4).

\bibitem[{Linsley et~al.(2018)Linsley, Kim, Veerabadran, Windolf, and Serre}]{LinsleyKVWS18}
Drew Linsley, Junkyung Kim, Vijay Veerabadran, Charles Windolf, and Thomas Serre. 2018.
\newblock \href {https://proceedings.neurips.cc/paper/2018/hash/ec8956637a99787bd197eacd77acce5e-Abstract.html} {Learning long-range spatial dependencies with horizontal gated recurrent units}.
\newblock In \emph{Advances in Neural Information Processing Systems 31: Annual Conference on Neural Information Processing Systems 2018, NeurIPS 2018, December 3-8, 2018, Montr{\'{e}}al, Canada}, pages 152--164.

\bibitem[{Ma et~al.(2023)Ma, Zhou, Kong, He, Gui, Neubig, May, and Zettlemoyer}]{MaZKHGNMZ23}
Xuezhe Ma, Chunting Zhou, Xiang Kong, Junxian He, Liangke Gui, Graham Neubig, Jonathan May, and Luke Zettlemoyer. 2023.
\newblock \href {https://openreview.net/pdf?id=qNLe3iq2El} {Mega: Moving average equipped gated attention}.
\newblock In \emph{The Eleventh International Conference on Learning Representations, {ICLR} 2023, Kigali, Rwanda, May 1-5, 2023}. OpenReview.net.

\bibitem[{Maas et~al.(2011)Maas, Daly, Pham, Huang, Ng, and Potts}]{MaasDPHNP11}
Andrew~L. Maas, Raymond~E. Daly, Peter~T. Pham, Dan Huang, Andrew~Y. Ng, and Christopher Potts. 2011.
\newblock \href {https://aclanthology.org/P11-1015/} {Learning word vectors for sentiment analysis}.
\newblock In \emph{The 49th Annual Meeting of the Association for Computational Linguistics: Human Language Technologies, Proceedings of the Conference, 19-24 June, 2011, Portland, Oregon, {USA}}, pages 142--150. The Association for Computer Linguistics.

\bibitem[{Merity et~al.(2017)Merity, Xiong, Bradbury, and Socher}]{MerityX0S17}
Stephen Merity, Caiming Xiong, James Bradbury, and Richard Socher. 2017.
\newblock \href {https://openreview.net/forum?id=Byj72udxe} {Pointer sentinel mixture models}.
\newblock In \emph{5th International Conference on Learning Representations, {ICLR} 2017, Toulon, France, April 24-26, 2017, Conference Track Proceedings}. OpenReview.net.

\bibitem[{Nangia and Bowman(2018)}]{NangiaB18}
Nikita Nangia and Samuel~R. Bowman. 2018.
\newblock \href {https://doi.org/10.18653/v1/n18-4013} {Listops: {A} diagnostic dataset for latent tree learning}.
\newblock In \emph{Proceedings of the 2018 Conference of the North American Chapter of the Association for Computational Linguistics, {NAACL-HLT} 2018, New Orleans, Louisiana, USA, June 2-4, 2018, Student Research Workshop}, pages 92--99. Association for Computational Linguistics.

\bibitem[{Ouyang et~al.(2022)Ouyang, Wu, Jiang, Almeida, Wainwright, Mishkin, Zhang, Agarwal, Slama, Ray, Schulman, Hilton, Kelton, Miller, Simens, Askell, Welinder, Christiano, Leike, and Lowe}]{Ouyang0JAWMZASR22}
Long Ouyang, Jeffrey Wu, Xu~Jiang, Diogo Almeida, Carroll~L. Wainwright, Pamela Mishkin, Chong Zhang, Sandhini Agarwal, Katarina Slama, Alex Ray, John Schulman, Jacob Hilton, Fraser Kelton, Luke Miller, Maddie Simens, Amanda Askell, Peter Welinder, Paul~F. Christiano, Jan Leike, and Ryan Lowe. 2022.
\newblock \href {http://papers.nips.cc/paper\_files/paper/2022/hash/b1efde53be364a73914f58805a001731-Abstract-Conference.html} {Training language models to follow instructions with human feedback}.
\newblock In \emph{Advances in Neural Information Processing Systems 35: Annual Conference on Neural Information Processing Systems 2022, NeurIPS 2022, New Orleans, LA, USA, November 28 - December 9, 2022}.

\bibitem[{Qi et~al.(2024{\natexlab{a}})Qi, Chen, Gao, Li, Liu, Wu, and Zhou}]{abs-2403-02628}
Biqing Qi, Xingquan Chen, Junqi Gao, Dong Li, Jianxing Liu, Ligang Wu, and Bowen Zhou. 2024{\natexlab{a}}.
\newblock \href {https://doi.org/10.48550/ARXIV.2403.02628} {Interactive continual learning: Fast and slow thinking}.
\newblock \emph{CoRR}, abs/2403.02628.

\bibitem[{Qi et~al.(2024{\natexlab{b}})Qi, Gao, Chen, Li, Liu, Wu, and Zhou}]{qi2024contrastive}
Biqing Qi, Junqi Gao, Xingquan Chen, Dong Li, Jianxing Liu, Ligang Wu, and Bowen Zhou. 2024{\natexlab{b}}.
\newblock Contrastive augmented graph2graph memory interaction for few shot continual learning.
\newblock \emph{arXiv preprint arXiv:2403.04140}.

\bibitem[{Qi et~al.(2024{\natexlab{c}})Qi, Gao, Liu, Wu, and Zhou}]{QiGLWZ24}
Biqing Qi, Junqi Gao, Jianxing Liu, Ligang Wu, and Bowen Zhou. 2024{\natexlab{c}}.
\newblock \href {https://doi.org/10.1109/LSP.2024.3383797} {Enhancing adversarial transferability via information bottleneck constraints}.
\newblock \emph{{IEEE} Signal Process. Lett.}, 31:1414--1418.

\bibitem[{Qi et~al.(2024{\natexlab{d}})Qi, Gao, Luo, Liu, Wu, and Zhou}]{abs-2402-16397}
Biqing Qi, Junqi Gao, Yiang Luo, Jianxing Liu, Ligang Wu, and Bowen Zhou. 2024{\natexlab{d}}.
\newblock \href {https://doi.org/10.48550/ARXIV.2402.16397} {Investigating deep watermark security: An adversarial transferability perspective}.
\newblock \emph{CoRR}, abs/2402.16397.

\bibitem[{Qi et~al.(2023)Qi, Zhang, Li, Tian, Zeng, Chen, and Zhou}]{qi2023large}
Biqing Qi, Kaiyan Zhang, Haoxiang Li, Kai Tian, Sihang Zeng, Zhang-Ren Chen, and Bowen Zhou. 2023.
\newblock \href {http://arxiv.org/abs/2311.05965} {Large language models are zero shot hypothesis proposers}.

\bibitem[{Qin et~al.(2023)Qin, Han, Sun, He, Li, Li, Dai, Kong, and Zhong}]{QinHSHLLDKZ23}
Zhen Qin, Xiaodong Han, Weixuan Sun, Bowen He, Dong Li, Dongxu Li, Yuchao Dai, Lingpeng Kong, and Yiran Zhong. 2023.
\newblock \href {https://openreview.net/pdf?id=IxmWsm4xrua} {Toeplitz neural network for sequence modeling}.
\newblock In \emph{The Eleventh International Conference on Learning Representations, {ICLR} 2023, Kigali, Rwanda, May 1-5, 2023}. OpenReview.net.

\bibitem[{Radev et~al.(2013)Radev, Muthukrishnan, Qazvinian, and Abu{-}Jbara}]{RadevMQA13}
Dragomir~R. Radev, Pradeep Muthukrishnan, Vahed Qazvinian, and Amjad Abu{-}Jbara. 2013.
\newblock \href {https://doi.org/10.1007/s10579-012-9211-2} {The {ACL} anthology network corpus}.
\newblock \emph{Lang. Resour. Evaluation}, 47(4):919--944.

\bibitem[{Schirmer et~al.(2022)Schirmer, Eltayeb, Lessmann, and Rudolph}]{SchirmerELR22}
Mona Schirmer, Mazin Eltayeb, Stefan Lessmann, and Maja Rudolph. 2022.
\newblock \href {https://proceedings.mlr.press/v162/schirmer22a.html} {Modeling irregular time series with continuous recurrent units}.
\newblock In \emph{International Conference on Machine Learning, {ICML} 2022, 17-23 July 2022, Baltimore, Maryland, {USA}}, volume 162 of \emph{Proceedings of Machine Learning Research}, pages 19388--19405. {PMLR}.

\bibitem[{Smith et~al.(2023)Smith, Warrington, and Linderman}]{SmithWL23}
Jimmy T.~H. Smith, Andrew Warrington, and Scott~W. Linderman. 2023.
\newblock \href {https://openreview.net/pdf?id=Ai8Hw3AXqks} {Simplified state space layers for sequence modeling}.
\newblock In \emph{The Eleventh International Conference on Learning Representations, {ICLR} 2023, Kigali, Rwanda, May 1-5, 2023}. OpenReview.net.

\bibitem[{Tabuada(2007)}]{Tabuada07}
Paulo Tabuada. 2007.
\newblock \href {https://doi.org/10.1109/TAC.2007.904277} {Event-triggered real-time scheduling of stabilizing control tasks}.
\newblock \emph{{IEEE} Trans. Autom. Control.}, 52(9):1680--1685.

\bibitem[{Tay et~al.(2021)Tay, Dehghani, Abnar, Shen, Bahri, Pham, Rao, Yang, Ruder, and Metzler}]{Tay0ASBPRYRM21}
Yi~Tay, Mostafa Dehghani, Samira Abnar, Yikang Shen, Dara Bahri, Philip Pham, Jinfeng Rao, Liu Yang, Sebastian Ruder, and Donald Metzler. 2021.
\newblock \href {https://openreview.net/forum?id=qVyeW-grC2k} {Long range arena : {A} benchmark for efficient transformers}.
\newblock In \emph{9th International Conference on Learning Representations, {ICLR} 2021, Virtual Event, Austria, May 3-7, 2021}. OpenReview.net.

\bibitem[{Vallarella and Haimovich(2019)}]{VallarellaH19}
Alexis~J. Vallarella and Hernan Haimovich. 2019.
\newblock \href {https://doi.org/10.1109/TAC.2018.2874669} {State measurement error-to-state stability results based on approximate discrete-time models}.
\newblock \emph{{IEEE} Trans. Autom. Control.}, 64(8):3308--3315.

\bibitem[{Vaswani et~al.(2017)Vaswani, Shazeer, Parmar, Uszkoreit, Jones, Gomez, Kaiser, and Polosukhin}]{vaswani2017attention}
Ashish Vaswani, Noam Shazeer, Niki Parmar, Jakob Uszkoreit, Llion Jones, Aidan~N Gomez, {\L}ukasz Kaiser, and Illia Polosukhin. 2017.
\newblock Attention is all you need.
\newblock \emph{Advances in neural information processing systems}, 30.

\bibitem[{Wang et~al.(2020)Wang, Li, Khabsa, Fang, and Ma}]{wang2020linformer}
Sinong Wang, Belinda~Z Li, Madian Khabsa, Han Fang, and Hao Ma. 2020.
\newblock Linformer: Self-attention with linear complexity.
\newblock \emph{arXiv preprint arXiv:2006.04768}.

\bibitem[{Zhu et~al.(2021)Zhu, Ping, Xiao, Shoeybi, Goldstein, Anandkumar, and Catanzaro}]{ZhuPXSGAC21}
Chen Zhu, Wei Ping, Chaowei Xiao, Mohammad Shoeybi, Tom Goldstein, Anima Anandkumar, and Bryan Catanzaro. 2021.
\newblock \href {https://proceedings.neurips.cc/paper/2021/hash/9425be43ba92c2b4454ca7bf602efad8-Abstract.html} {Long-short transformer: Efficient transformers for language and vision}.
\newblock In \emph{Advances in Neural Information Processing Systems 34: Annual Conference on Neural Information Processing Systems 2021, NeurIPS 2021, December 6-14, 2021, virtual}, pages 17723--17736.

\bibitem[{Zuo et~al.(2022)Zuo, Liu, Jiao, Charles, Manavoglu, Zhao, and Gao}]{abs-2212-08136}
Simiao Zuo, Xiaodong Liu, Jian Jiao, Denis Charles, Eren Manavoglu, Tuo Zhao, and Jianfeng Gao. 2022.
\newblock \href {https://doi.org/10.48550/ARXIV.2212.08136} {Efficient long sequence modeling via state space augmented transformer}.
\newblock \emph{CoRR}, abs/2212.08136.

\end{thebibliography}

\clearpage
\appendix
\onecolumn
\section{Appendix}
\subsection{Proofs}
\paragraph{Proof of Proposition 1}
Denote the sampled $u'_t = u_t+\varepsilon_t$, where $\varepsilon_t$ is the sampling error caused by variation in the sampling points. Consider the propagation of the error in the output values $\{y_k\}_{k=1}^L$:
\begin{equation}
\begin{bmatrix}
 y'_1\\
 y'_2\\
 \vdots\\
 y'_L
\end{bmatrix} = 
\begin{bmatrix}
 \overline{\boldsymbol {CB}} & \mathbf 0 & \cdots & \mathbf 0 \\
 \overline{\boldsymbol {CAB}} & \overline{\boldsymbol {CB}} & \cdots &  \mathbf 0\\
 \vdots & \vdots & \ddots  & \vdots \\
 \overline{\boldsymbol {CA}}^{T-1}\overline{\boldsymbol {B}} & \overline{\boldsymbol {CA}}^{T-2}\overline{\boldsymbol {B}} & \cdots & \overline{\boldsymbol {CB}}
\end{bmatrix}\begin{bmatrix}
 u_1+\varepsilon_1\\
 u_2+\varepsilon_2\\
 \vdots\\
 u_T+\varepsilon_t
\end{bmatrix},
\end{equation}
then
\begin{equation}
\begin{aligned}
\|y'_t-y_t\|&= \left\|\overline{\boldsymbol {CA}}^{t-1}\overline{\boldsymbol {B}}\varepsilon_1+\overline{\boldsymbol {CA}}^{t-2}\overline{\boldsymbol {B}}\varepsilon_2+\dots+\overline{\boldsymbol {CB}}\varepsilon_t\right\|\\
& \le \left\|\overline{\boldsymbol C}\right\|\left\|\overline{\boldsymbol {A}}^{t-1}\right\|\left\|\overline{\boldsymbol {B}}\right\|\left|\varepsilon_1\right|+\left\|\overline{\boldsymbol C}\right\|\left\|\overline{\boldsymbol {A}}^{t-2}\right\|\left\|\overline{\boldsymbol {B}}\right\|\left|\varepsilon_2\right|+\dots+\left\|\overline{\boldsymbol C}\right\|\left\|\overline{\boldsymbol {B}}\right\|\left|\varepsilon_t\right|\\
& \le \left|\lambda_{\max}\right|^{t-1}cb\varepsilon_1 + \left|\lambda_{\max}\right|^{t-2}cb\varepsilon_2 + \dots +cb\varepsilon_t.
\end{aligned}
\end{equation}

Note that if $\lambda_{\max} \geq 1$, $\lim_{t\rightarrow \infty}\|y'_t-y_t\|$ becomes unbounded. If $\left|\lambda_{\max}\right|<1$, then we have
\begin{equation}
\begin{aligned}
\|\boldsymbol{x}_t\|&=\left\|\overline{\boldsymbol {A}}^{L-1}\overline{\boldsymbol {B}}u_1+\overline{\boldsymbol {A}}^{L-2}\overline{\boldsymbol {B}}u_2+\dots+\overline{\boldsymbol {B}}u_t\right\| \\
&\le \left\|\overline{\boldsymbol {A}}^{L-1}\right\|\left\|\overline{\boldsymbol {B}}\right\|\left|u_1\right|+\left\|\overline{\boldsymbol {A}}^{L-2}\right\|\left\|\overline{\boldsymbol {B}}\right\|\left|u_2\right|+\dots+\left\|\overline{\boldsymbol {B}}\right\|\left|u_t\right|\\
&\le \left|\lambda_{\max}\right|^{L-1}b\zeta + \left|\lambda_{\max}\right|^{L-2}b\zeta + \dots +b\zeta,
\end{aligned}
\end{equation}
thus
\begin{equation}
\lim_{t\rightarrow\infty}\|\boldsymbol{x}_t\| \le \lim_{t\rightarrow \infty}\left( \left|\lambda_{\max}\right|^{L-1}b\zeta + \left|\lambda_{\max}\right|^{L-2}b\zeta + \dots +b\zeta\right)=\frac{b\zeta}{1-\left|\lambda_{\max}\right|}<\lim_{t\rightarrow\infty}\|\boldsymbol{x}_t\|,
\end{equation}
which contradicts the assumption, therefore there must be $\left|\lambda_{\max}\right|>=1$, which also implies that $\lim_{t\rightarrow \infty}\|y'_t-y_t\|$ is unbounded.

\textbf{Remark} 
Note that imposing the constraint $\left|\lambda_{\max}\right|<1$ on the state space model will cause the initial input $u_{t_0}$ to tend to zero as it propagates ($\overline{\boldsymbol {A}}^{t-t_0}\overline{\boldsymbol {B}}u_{t_0} \underset{t-t_0\rightarrow \infty}{\longrightarrow} 0$). This causes all previous states to rapidly decay to $0$ during the propagation, thus severely limits the long-term memory capacity of the model.

\paragraph{Proof of Theorem1}
Taking into account the error propagation in latent states of the SSM model, the grid deviation error emerges from signal misalignment and can be considered as an additional disturbance term. Assuming that the actual sampled value, denoted as $u'$, satisfies the relationship $u'_t = u_t+\varepsilon_t$, where $\varepsilon_t$ represents the error term, we can have
\begin{equation}
\begin{bmatrix}
 \boldsymbol{x}_1\\
 \boldsymbol{x}_2\\
 \vdots\\
 \boldsymbol{x}_T
\end{bmatrix} = 
\begin{bmatrix}
 \overline{\boldsymbol B} & \mathbf 0 & \cdots & \mathbf 0 \\
 \overline{\boldsymbol {AB}} & \overline{\boldsymbol B} & \cdots &  \mathbf 0\\
 \vdots & \vdots & \ddots  & \vdots \\
 \overline{\boldsymbol {A}}^{T-1}\overline{\boldsymbol {B}} & \overline{\boldsymbol {A}}^{T-2}\overline{\boldsymbol {B}} & \cdots & \overline{\boldsymbol {B}}
\end{bmatrix}\begin{bmatrix}
 u_1+\varepsilon_1\\
 u_2+\varepsilon_2\\
 \vdots\\
 u_T+\varepsilon_t
\end{bmatrix},
\end{equation}
observe that
\begin{equation}
\begin{aligned}
\boldsymbol{x}_t &= \overline{\boldsymbol {A}}^{t-1}\overline{\boldsymbol {B}}(u_1+\varepsilon_1) + \overline{\boldsymbol {A}}^{t-2}\overline{\boldsymbol {B}}(u_2+\varepsilon_2) + \dots + \overline{\boldsymbol {B}}(u_t+\varepsilon_t)\\
 &= \overline{\boldsymbol {A}}^{t-1}\overline{\boldsymbol {B}}u_1 + \overline{\boldsymbol {A}}^{t-2}\overline{\boldsymbol {B}}u_2 + \dots + \overline{\boldsymbol {B}}u_t + L(\varepsilon_1,\varepsilon_2,\dots,\varepsilon_t),
\end{aligned}
\end{equation}
where $L(\varepsilon_1,\varepsilon_2,\dots,\varepsilon_t) = \overline{\boldsymbol {A}}^{t-1}\overline{\boldsymbol {B}}\varepsilon_1 + \overline{\boldsymbol {A}}^{t-2}\overline{\boldsymbol {B}}\varepsilon_2 + \dots + \overline{\boldsymbol {B}}\varepsilon_t$.
%Therefore, we consider the following form of state propagation (controller):
Consider its continuous form and drawing upon the controller concept in ETC theory, we consider the following state propagation:
\begin{equation}
\dot {\boldsymbol{x}}(t) = \boldsymbol{A}\left(\boldsymbol{x}(t)+\int_{0}^t \boldsymbol{k}(t-l)\varepsilon(l)dl\right)+  \boldsymbol{B}u(t), 
\end{equation}
where $\boldsymbol{k}$ is a coefficient matrix that varies over time, and has the same shape as $\overline{\boldsymbol{B}}$.

Owing to the accumulation of errors in the time domain, we introduce a modifiable factor denoted as $h([t-\tau,t])$ with backtracking capability to regulate the input. Specifically, the controlled input is defined as $u_{adj}(t)=h([l-\tau,l])u(t)$.
then we have
\begin{equation}
\dot{\boldsymbol{x}}(t) = \boldsymbol{A}\left(x(t)+\int_{0}^t \boldsymbol{k}(t-l)h([l-\tau,l])\varepsilon(l)dl\right)+  \boldsymbol{B}h([t-\tau,t])u(t), 
\end{equation}
then $h_\tau(t)$ has the ability to adjust the errors with coefficients carrying temporal phases. Taking into account the following observer used for sampling:
\begin{equation}
\dot {\boldsymbol{z}}(t)=\boldsymbol A\boldsymbol{z}(t) + \boldsymbol Bh([t-\tau,t])(u(t)+\varepsilon(t)),
\end{equation}
denote $\boldsymbol{e}(t)=\boldsymbol{x}(t)-\boldsymbol{z}(t)$, we have
\begin{equation}
\dot{\boldsymbol e}(t) = \boldsymbol{A}\boldsymbol{e}(t)+\boldsymbol{A}\int_0^t\boldsymbol{k}(t-l)h([l-\tau,l])\varepsilon(l)dl-\boldsymbol{B}h([t-\tau,t])\varepsilon(t).
\end{equation}
Consider the Lyapunov function $\mathcal{L}_{\boldsymbol{e}}(t)=\boldsymbol{e}^\top(t)\boldsymbol{P}\boldsymbol{e}(t)$, where $\boldsymbol{P}$ is a positive definite symmetric matrix, we can obtain
\begin{equation}
\begin{aligned}
\frac{d\mathcal{L}_{\boldsymbol{e}}(t)}{dt} & = 2\boldsymbol{e}^\top(t)\boldsymbol{P}\dot{\boldsymbol{e}}(t) \\
&=2\boldsymbol{e}^\top(t)\boldsymbol{P}\left(\boldsymbol{A}\boldsymbol{e}(t)+\boldsymbol{A}\int_0^t\boldsymbol{k}(t-l)h([l-\tau,l])\varepsilon(l)dl-\boldsymbol{B}h([t-\tau,t])\varepsilon(t)\right)\\
&=\boldsymbol{e}^\top(t)\left(\boldsymbol{PA}+\boldsymbol{A^\top P}\right)\boldsymbol{e}(t)+ \Lambda (t),
\end{aligned}
\end{equation}

where 
\begin{equation}
\begin{aligned}
\Lambda (t)&=2\boldsymbol{e}^\top(t)\boldsymbol{A}\int_0^t\boldsymbol{k}(t-l)h([l-\tau,l])\varepsilon(l)dl-\boldsymbol{e}^\top(t)\boldsymbol{B}h([t-\tau,t])\varepsilon(t)\\
&=2\left\|\boldsymbol{e}(t)\right\|\left\|\boldsymbol{A}\right\|\int_0^t\left\|\boldsymbol{k}(t-l)\right\|\left|h([l-\tau, l])\right|\left|\varepsilon(l)\right|dl +\left\|\boldsymbol{e}(t)\right\|\left\|\boldsymbol{B}\right\|\left|h([t-\tau, t])\right|\left|\varepsilon(t)\right| \\
&\le 2\left\|h_{\tau}\right\|\left\|\boldsymbol{e}(t)\right\|\left(\int_0^t\left\|\boldsymbol{k}(t-l)\right\|\left|\varepsilon(l)\right|dl+\left\|\boldsymbol{B}\right\|\left|\varepsilon(t)\right|\right).
\end{aligned}
\end{equation}
Hence, selecting a value of $\left|h([t-\tau,t])\right|<1$ strengthens the stability of the system, while $h([t-\tau,t])\equiv 1$ corresponds to the case without a controller. Additionally, choosing a larger $\tau$ value can further enhance the control performance.
\twocolumn
\clearpage
\begin{figure*}[t]
\centering  
\subfigure{
\label{fig8.1}
\includegraphics[width=0.233\textwidth]{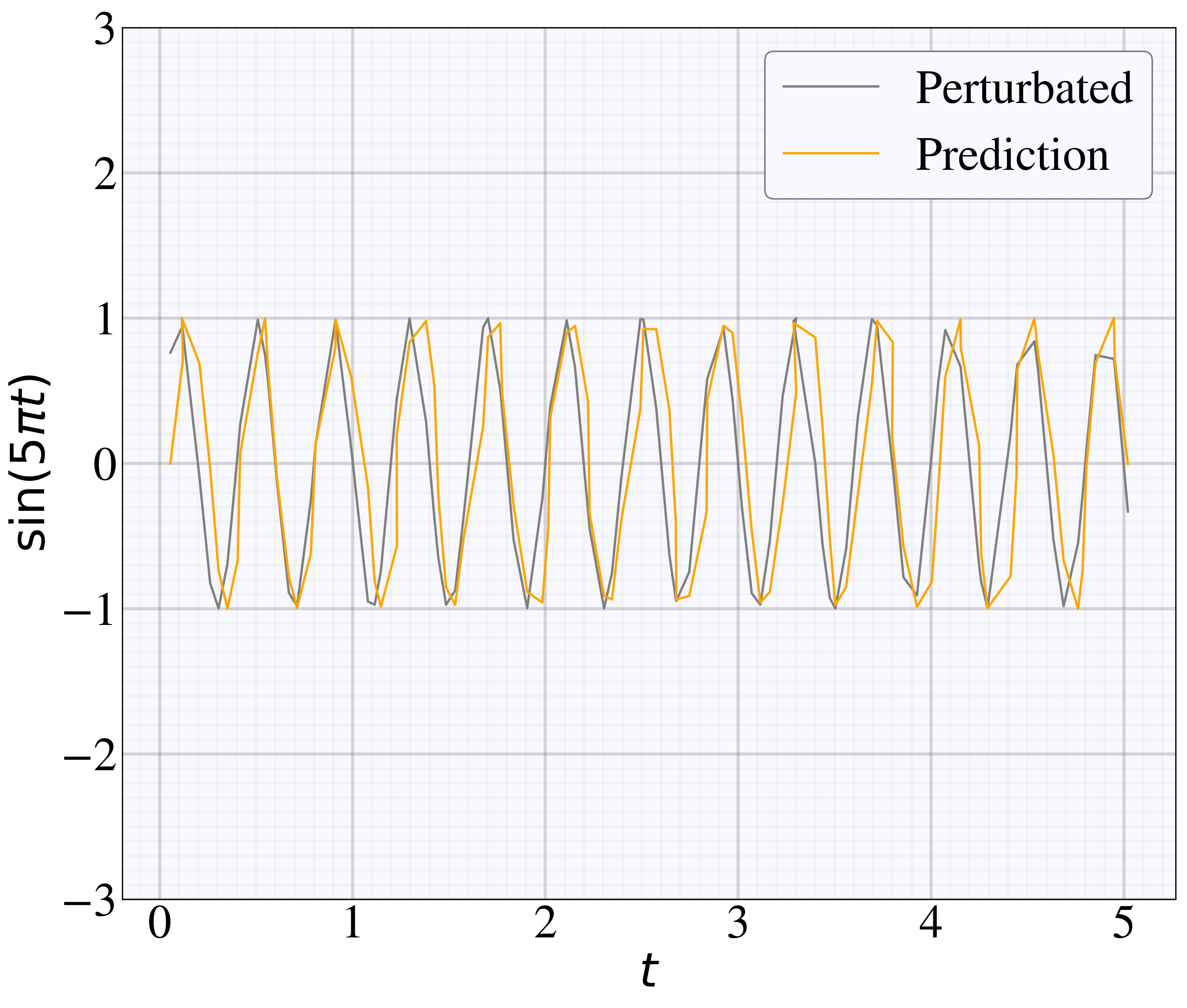}}
\subfigure{
\label{fig8.2}
\includegraphics[width=0.233\textwidth]{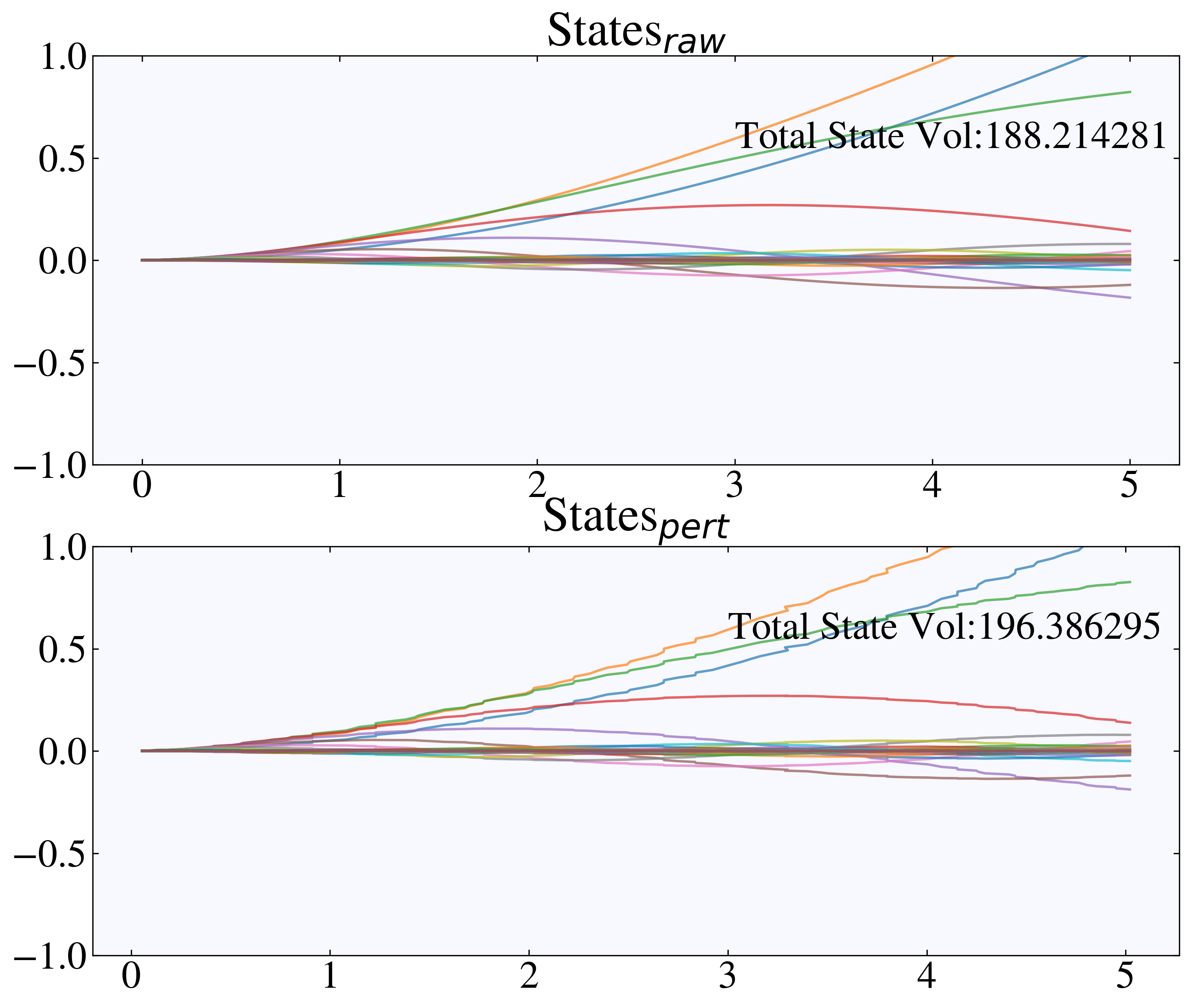}}
\subfigure{
\label{fig8.3}
\includegraphics[width=0.233\textwidth]{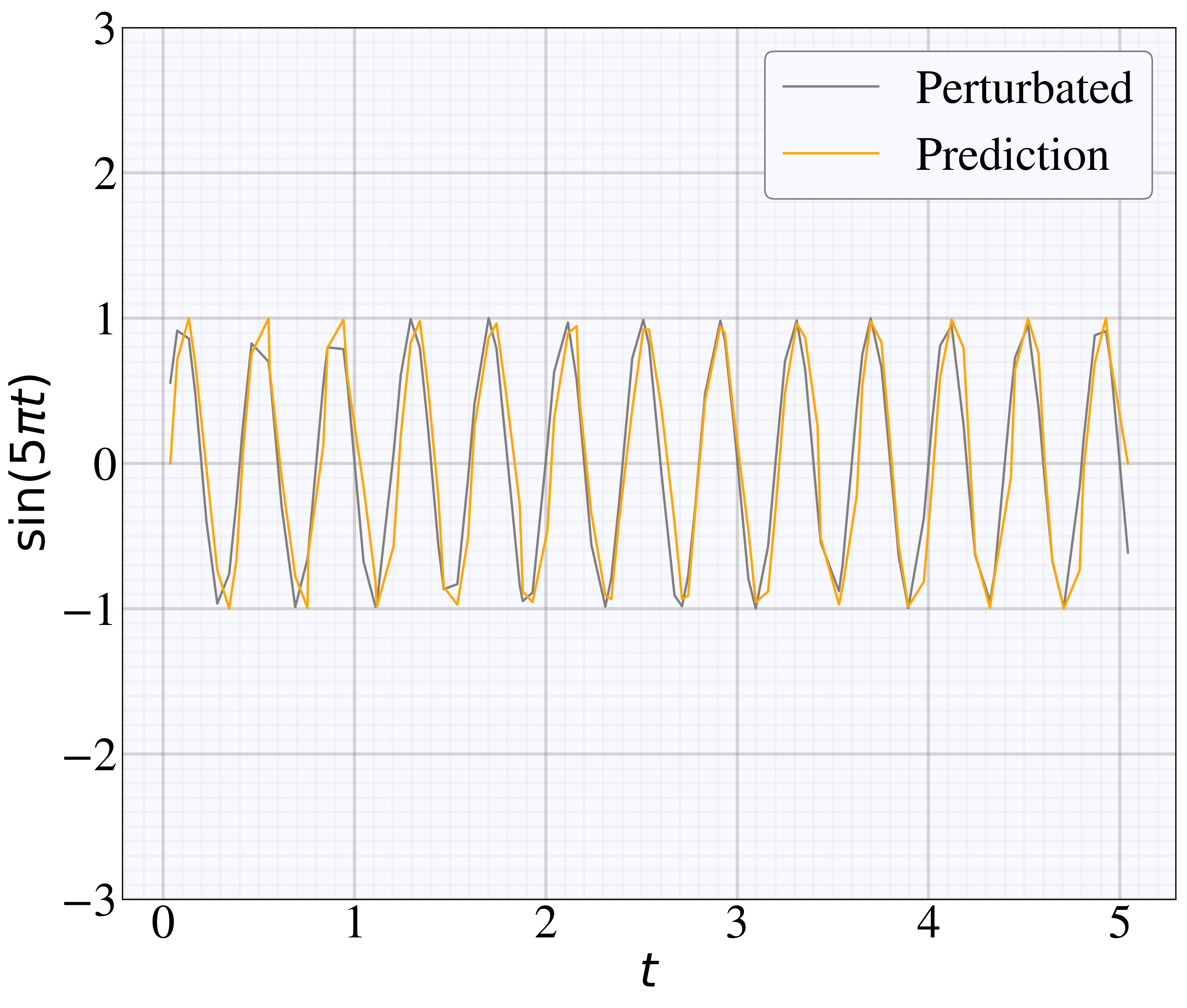}}
\subfigure{
\label{fig8.4}
\includegraphics[width=0.233\textwidth]{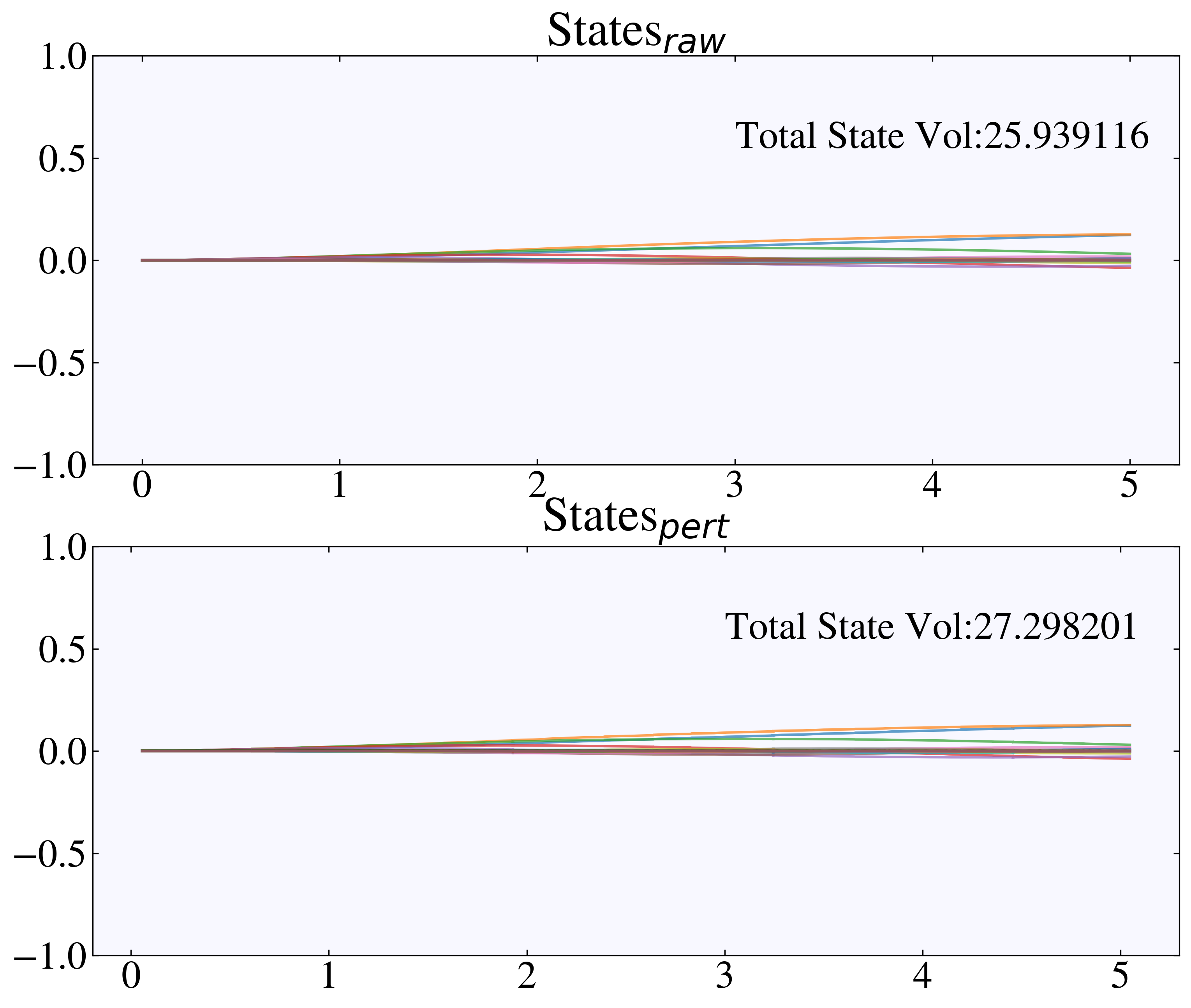}}
\vspace{-5pt}
\caption{Comparative results for w/ and w/o SMR in 5-layer S4, the incorporation of SMR alleviate the NSS problem.}
\label{Figure 8}

\end{figure*}

\begin{figure*}[t]
\centering  
\includegraphics[width=1\textwidth]{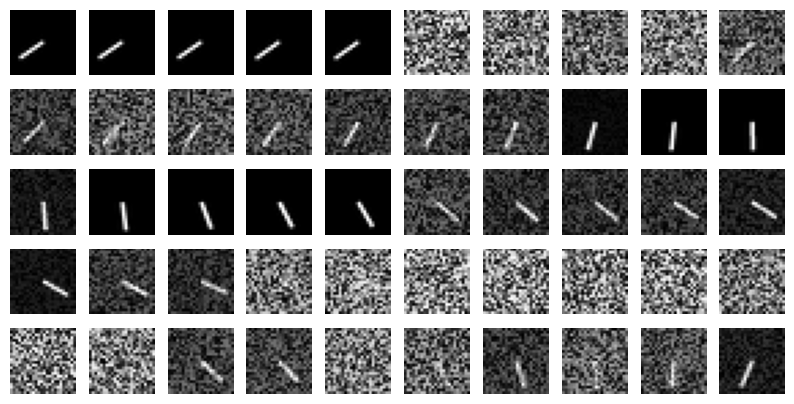}
\caption{Input example of the used Pendulum dataset.}
\label{fig.9}

\end{figure*}

\subsection{NSS in 5-layers S4}
\label{supp.B}
Due to space constraints, we present the analysis of the deep S4 model here. Specifically, we conducted an experiment on a 5-layer S4 model, extending from the experiment described in Section 2.3. We plotted the results of the hidden states in the first layer and observed the presence of the NSS issue in the 5-layer S4 model, as depicted in Fig.\ref{fig8.2}. Notably, the S4 model without SMR exhibited a significant NSS phenomenon. In contrast, the S4 model incorporated with SMR demonstrated highly stable hidden states, as illustrated in Fig.\ref{fig8.4}. The sum of absolute values of the states at each time step decreased from $10^2$ to $10^1$, and the output error under perturbation was also reduced (Fig.\ref{Figure 8}).

\subsection{Example of Pendulum Dataset}
\label{Appendix C}
We present the input examples of the pendulum dataset used in Section \ref{2.4} in Fig.\ref{fig.9}. The sampling intervals are not constant but variable, and the introduction of random noise in the image sequence makes the actual sampling intervals even more random. All models are uniformly adjusted to $4$ blocks with a hidden dimension of $64$, and optimized using the AdamW optimizer with a learning rate of $1e-4$.
\begin{table}[t]
\centering
\scriptsize
\caption{Detailed training settings used in our experiments.}
\label{tab_wiki}
\setlength{\tabcolsep}{12pt}
\renewcommand{\arraystretch}{1.2}
\begin{tabular}{c|c}
\hline\hline
 & \textbf{Autoregressive language modelling} \\
    \hline\hline
    Data used & Wikitext-103  \\
    Tokenizer method & BPE   \\
    Vocab size & 50265  \\
    Sequence length & 512  \\
    Batch size & 64 \\
    Total updates & 50,000 \\
    Warmup steps & 3,000 \\
    Peak learning rate & 5e-4\\
    Lr scheduler & Inverse sqrt\\
    Optimizer & Adam \\
    Adam $\epsilon$ & 1e-8 \\
    Adam $(\beta_1, \beta_2)$ & (0.9, 0.98) \\
    Weight decay & 0.1 \\
    Gradient clip norm & 1.0 \\
    Dropout & 0.1  \\
    \hline\hline
  \end{tabular}
\end{table}
\subsection{Experiment Details}
\label{Appendix D}
Here, we provide specific configurations for the experiments mentioned in Section \ref{4}. The experimental settings for autoregressive language modeling are detailed in Tab.\ref{tab_wiki}, while the parameter configurations for various tasks on the LRA are presented in Tab.\ref{tab_lra}.

\begin{table*}[t]
\centering
\small
\caption{Detailed training settings used in LRA tasks.}
\label{tab_lra}
\setlength{\tabcolsep}{12pt}
\renewcommand{\arraystretch}{1.2}
\begin{tabular}{c||c|c|c|c|c}
\hline\hline
 & Retrieval & ListOps & Text & Image & Pathfinder  \\
    \hline\hline
    Num blocks & 6 & 6 & 4 & 6 & 4\\
    Embedding dimension & 256 & 128 & 128 & 512 & 128\\
    Max length & 4000 & 2048 & 4096 & 1024 & 1024  \\
    Batch size & 16 & 50 & 50 & 50 & 64\\
    Total epochs & 20 & 40 & 50 & 200 & 200 \\
    Learning rate & 1e-3 & 3e-3 & 1e-3 & 4e-3 & 4e-3 \\
    Weight decay & 0.0 & 0.04 & 5e-2 & 3e-2 & 3e-2 \\
    Dropout & 0.0 & 0.0 & 0.1 & 0.1 & 0.1  \\
    \hline\hline
  \end{tabular}
\end{table*}

\end{document}